%% file: NeLo.tex
\documentclass[acmtog,screen,nonacm]{acmart}
\acmSubmissionID{292}

\usepackage{booktabs}
\usepackage{multirow}
\usepackage{epsfig}
\usepackage{graphicx}
\usepackage{amsmath}
\usepackage{amssymb}
\usepackage{overpic}
\usepackage{enumitem}
\usepackage{layouts}
\usepackage{hyphenat,balance,microtype}
\usepackage[fleqn,tbtags]{mathtools}
\usepackage[capitalise]{cleveref}
\usepackage[detect-weight]{siunitx}
\usepackage{caption, subcaption, textpos}
\usepackage{colortbl}

\usepackage{algorithm}
\usepackage{algorithmic}
\usepackage{verbatim}
\usepackage{listings}
\usepackage{wrapfig}

\graphicspath{{./images/}}

\definecolor{Highlight}{HTML}{00cc98} 
\newcommand{\hl}[1]{\textcolor{Highlight}{#1}}

\newcommand{\rev}[1]{\textcolor{black}{#1}}

\newcommand{\tablestyle}[2]{\setlength{\tabcolsep}{#1}
                            \renewcommand{\arraystretch}{#2}
                            \centering
                            \footnotesize}

\acmJournal{TOG}
\acmYear{2024}
\acmVolume{43}
\acmNumber{6}
\acmArticle{0}
\acmMonth{12}

\setcopyright{acmcopyright}

\acmDOI{10.1145/3687901}

\received{May 2024}
\received[accepted]{July 2024}
\received[final version]{September 2024}

\title{Neural Laplacian Operator for 3D Point Clouds}

\author{Bo Pang}
\affiliation{
  \institution{Peking University}
  \country{China}
}
\email{bo98@stu.pku.edu.cn}

\author{Zhongtian Zheng}
\affiliation{
  \institution{Peking University}
  \country{China}
}
\email{zhengzhongtian@pku.edu.cn}

\author{Yilong Li}
\affiliation{
  \institution{Peking University}
  \country{China}
}
\email{2201111647@pku.edu.cn}

\author{Guoping Wang}
\affiliation{
  \institution{Peking University}
  \country{China}
}
\email{wgp@pku.edu.cn}
\authornote{Corresponding authors. Peng-Shuai Wang is the project leader.}

\author{Peng-Shuai Wang}
\affiliation{
  \department{Wangxuan Institute of Computer Technology}
  \institution{Peking University}
  \country{China}
}
\email{wangps@hotmail.com}
\authornotemark[1]

\begin{abstract}
The discrete Laplacian operator holds a crucial role in 3D geometry processing, yet it is still challenging to define it on point clouds.
Previous works mainly focused on constructing a local triangulation around each point to approximate the underlying manifold for defining the Laplacian operator, which may not be robust or accurate.
In contrast, we simply use the $K$-nearest neighbors (KNN) graph constructed from the input point cloud and learn the Laplacian operator on the KNN graph with graph neural networks (GNNs).
However, the ground-truth Laplacian operator is defined on a manifold mesh with a different connectivity from the KNN graph and thus cannot be directly used for training.
To train the GNN, we propose a novel training scheme by imitating the behavior of the ground-truth Laplacian operator on a set of probe functions so that the learned Laplacian operator behaves similarly to the ground-truth Laplacian operator.
We train our network on a subset of ShapeNet and evaluate it across a variety of point clouds.
Compared with previous methods, our method reduces the error by \emph{an order of magnitude} and excels in handling sparse point clouds with thin structures or sharp features.
Our method also demonstrates a strong generalization ability to unseen shapes.
With our learned Laplacian operator, we further apply a series of Laplacian-based geometry processing algorithms directly to point clouds and achieve accurate results, enabling many exciting possibilities for geometry processing on point clouds.
The code and trained models are available at \url{https://github.com/IntelligentGeometry/NeLo}.
\end{abstract}

\begin{teaserfigure}
  \centering
  \includegraphics[width=0.98\linewidth]{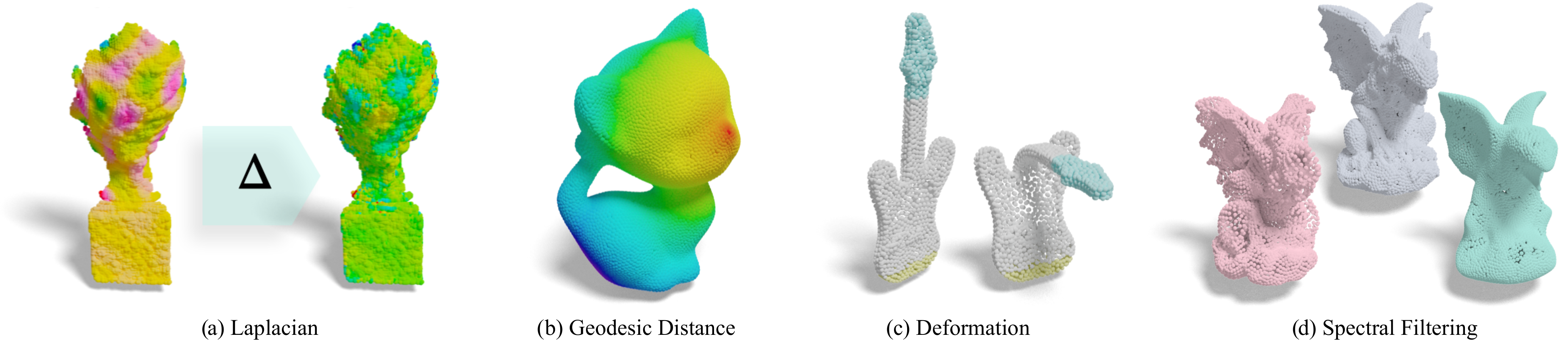}
  \caption{
    We propose to learn a Laplacian operator for point clouds with graph neural networks.
    Our neural Laplacian operator is accurate and generalizable, enabling many exciting possibilities for geometry processing on point clouds.
    Our neural Laplacian operator can be applied to a real world point cloud and compute the Laplacian of an arbitrary function (a).
    With our neural Laplacian operator, we can compute the geodesic distance on the point cloud (b), manipulate the point cloud with the ARAP deformation method (c), and perform spectral filtering on the point cloud (d).
    In (c), the colored area indicates the handles of the point cloud.
    In (d), the input point cloud is in gray, and the exaggerated and smoothed results are in magenta and cyan.
  }
  \label{fig:teaser}
\end{teaserfigure}



\begin{document}

\maketitle
\input{src/introduction}
\input{src/relatedwork}
\input{src/method}

\input{src/results}

\input{src/conclusion}

\begin{acks}
We thank the anonymous reviewers for their invaluable suggestions.
This work is supported by the National Key R\&D Program of China (No. 2022YFB3303400 and No. 2021YFF0500901), Shenzhen Science and Technology Program KJZD20230923114114028, and Beijing Natural Science Foundation No. 4244081.
\end{acks}

\bibliographystyle{ACM-Reference-Format}
\bibliography{src/reference}


\end{document}

%% file: src/introduction.tex
\section{Introduction} \label{sec:intro}

The Laplacian operator plays a central role in graphics and 3D geometry processing~\cite{Botsch2010} and is widely used in shape analysis~\cite{Sun2009a}, parameterization~\cite{Floater2005}, smoothing~\cite{Desbrun1999}, editing~\cite{Yu2004}, spectral processing~\cite{Vallet2008}, and more.
With the advancements of 3D acquisition techniques, 3D point clouds are now ubiquitous and have become a popular representation of 3D geometry.
Although the Laplacian operator is well defined on manifold meshes~\cite{Meyer2003,Desbrun1999,Pinkall1993}, it is still an open problem to define the Laplacian operator on point clouds~\cite{Sharp2020a}, which makes it difficult or even impossible to \emph{directly} apply Laplacian-based geometry processing methods on point clouds.

The key problem is that point clouds lack well-defined manifold structures, which is a prerequisite for the Laplacian operator since the Laplacian operator is essentially a differential operator that measures the local variation of a function on a manifold.
A common approach to defining the Laplacian operator on point clouds involves constructing a local triangulation using the $K$-nearest neighbors around each point to approximate the local manifold structure, then building the Laplacian operator with the triangulation in the estimated tangent plane~\cite{Belkin2009,Luo2009,Liu2012,Qin2018,Sharp2020a} or in 3D space~\cite{Clarenz2004,Cao2010}.
However, the task of triangulation is non-trivial, and the local triangulation may be inconsistent with the underlying ground-truth manifold surfaces, especially when the point clouds contain noise, sharp features, or thin structures.
Moreover, these methods may have multiple parameters that require manual tuning, severely restricting their practicality.
As a result, current Laplacian operators on point clouds are not robust and are inaccurate in many cases, thus far from satisfactory.

Mathematically, the Laplacian operator for a point cloud is a linear operator and can be represented by a matrix.
This matrix is known as the Laplacian matrix and can be depicted as a graph, where each point is a node, and the non-zero matrix elements represent edges.
Therefore, two key factors of the Laplacian operator are the graph structure and the edge weights.
Previous studies concentrated on the graph structure, aiming to construct a well-defined local triangulation for each point with heuristic strategies.
These efforts strive to make the local triangulation Delaunay-like~\cite{Belkin2009,Liu2012} and as manifold as possible~\cite{Sharp2020a}, which turns out to be extremely hard.
Counterintuitively, we diverge from this path and abandon the attempts to enhance the quality of the graph structure.
We opt to directly use the KNN graph trivially constructed from the input point cloud, although the KNN graph is not even a valid triangulation. 
Our goal is to derive appropriate edge weights for the KNN graph, so that the Laplacian operator defined on the KNN graph closely approximates the ground-truth Laplacian operator on the underlying surfaces.
Of course, the edge weights cannot be manually determined explicitly.
To address this, our key idea is to learn the edge weights with graph neural networks (GNNs) from data.
Given a KNN graph constructed from a point cloud, we use a GNN to process the graph and extract a high-dimensional feature for each point.
Subsequently, we employ a multi-layer perceptron (MLP) to project the squared difference of features between neighboring points to the edge weights and append a ReLU activation function to ensure the edge weights are positive.
The learned edge weights are then utilized to assemble the Laplacian matrix.
Additionally, we use another MLP to project each point feature to a scalar, simulating the effect of the mass matrix.
As we learn the Laplacian operator with a neural network, we term it the \emph{neural Laplacian operator}, abbreviated as \textsc{NeLo}.

However, the training of such a GNN is a nontrivial task.
Our neural Laplacian operator is defined on a KNN graph, whereas the ground-truth Laplacian operators are defined on the corresponding manifold mesh.
The connectivities of the KNN graph and the ground-truth mesh are totally different; thus, we do not have ground-truth edge weights to supervise the training of the GNN.
Inspired by the idiom ``\textit{If it looks like a duck, swims like a duck, quacks like a duck, then it probably is a duck}'', our key insight is that we do not need to replicate the ground-truth Laplacian operator itself, but only need to imitate its behavior.
We assume that if the neural Laplacian operator behaves very closely to the ground-truth Laplacian operator, then in practice, they could be considered the same operator.
Specifically, we select a series of probe functions defined on the 3D shapes and let both the ground-truth Laplacian operator and the neural Laplacian operator process these functions. 
Then, we train the GNN by minimizing the difference between the results of the two Laplacian operators.
In other words, we do not try to enforce the neural Laplacian operator identical to the ground-truth Laplacian operator, but instead train the GNN to make them ``behave'' similarly.

For the design of probe functions, theoretically, all discrete functions defined on the point cloud can be expanded as a linear combination of the eigenvectors of the ground-truth Laplacian matrix.
Ideally, we could use all eigenvectors of the ground-truth Laplacian matrix as the probe functions, which naturally cover all possible functions defined on the point cloud.
However, computing all eigenvectors is computationally expensive.
We thus opt to select the top-k low-frequency eigenvectors as the probe functions, which prove to be sufficient to represent most functions defined on the point cloud, and for downstream applications.
To ensure the learned Laplacian operator is accurate for high-frequency functions, we additionally incorporate a set of trigonometric functions with random frequencies. \looseness=-1

We trained the GNN to predict the Laplacian operator using the proposed probing functions on approximately $9.4k$ point clouds from a subset of ShapeNet~\cite{Chang2015}.
The experimental results demonstrate that our neural Laplacian operator is capable of approximating the ground-truth Laplacian operator across a diverse range of point clouds, significantly outperforming other state-of-the-art methods~\cite{Sharp2020a, Belkin2009}.
Our neural Laplacian operator also exhibits strong robustness to noise and displays a strong ability to generalize to unseen shapes.
More importantly, our neural Laplacian operator empowers the straightforward application of numerous Laplacian-based geometry processing algorithms to point clouds, such as heat diffusion, geodesic distance~\cite{Crane2013}, smoothing~\cite{Desbrun1999}, deformation~\cite{Sorkine2007}, and spectral filtering~\cite{Vallet2008}.
Our results closely align with the ground-truth results on the corresponding meshes.
We expect that our neural Laplacian operator may inspire more applications of Laplacian-based geometry processing algorithms on point clouds.

In conclusion, our contributions are as follows:
\begin{itemize}[leftmargin=16pt,itemsep=2pt]
    \item[-] We propose a novel method to learn the Laplacian operator for point clouds on the KNN graphs with GNNs.
    \item[-] We propose a novel training scheme by imitating the behavior of the ground-truth Laplacian operator on a set of probe functions, which may also be applicable to other differential operators.
    \item[-] Our neural Laplacian operator is accurate, robust, and significantly better than other manually designed Laplacian operators, enabling Laplacian-based geometry processing algorithms to be directly applied to point clouds for accurate results.
\end{itemize}

%% file: src/relatedwork.tex
\section{Related Work} \label{sec:related}


\paragraph{Laplacian for Geometry Processing}
The discrete Laplace operator is one of the most important differential operators in 3D geometry processing~\cite{Botsch2010,Levy2006}.
The Laplacian operator is the key for a series of differential equations on 3D shapes, such as the Poisson equation, diffusion equation, and wave equation.
By solving the heat equation and Poisson equation, the geodesic distance on 3D shapes can be efficiently computed~\cite{Crane2013a,Crane2017}.
By simulating the melting process of 3D shapes with the heat equation, geometric filters can be developed for 3D shape denoising and smoothing~\cite{Taubin1995,Desbrun1999}.
Spectral filtering can be achieved by manipulating the eigenfunctions of the Laplacian operator to amplify or filter out specific frequency components~\cite{Vallet2008,Zhang2010}.
\rev{In addition, Keros and Subr~\shortcite{Keros2023} propose a spectral preserving coarsening algorithm for simplicial complexes.}
The Laplacian operators are also widely used in deformation and simulation.
Representative algorithms include the Poisson deformation~\cite{Yu2004} and the as-rigid-as-possible (ARAP) deformation~\cite{Sorkine2007}.
Recently, the Laplacian operator is also used in deep learning, for \rev{3D shape analysis~\cite{Sharp2022,Yi2017b,Dong2023}} and shape recovery~\cite{Marin2021}.
In our experiments, we apply some of these algorithms to point clouds to demonstrate the effectiveness of our neural Laplacian operator.

\paragraph{Laplacian Operator on Meshes}
The study of defining the Laplacian operator on manifold triangle meshes while considering properties like symmetry, local support, and semi-definiteness has spanned decades~\cite{Wardetzky2007}.
The simplest form of the Laplace operator is the graph Laplacian with uniform edge weights~\cite{Taubin1995}.
While the uniform Laplacian is easy to compute, it falls short in capturing the local geometry of the mesh.
The cotangent Laplacian operator takes into account the local geometry and provides a more precise approximation of the Laplacian operator on curved surfaces~\cite{Desbrun1999,Pinkall1993,Meyer2003}.
\rev{Subsequent works further extend the Laplacian operator to general polygonal meshes~\cite{Herholz2015,Alexa2011,De2020,Bunge2024}, or try to infer Laplacian-Beltrami operators in a data driven approach~\cite{Chazal2016}.}
Recently, \cite{Sharp2020a} introduced a Laplacian tailored for nonmanifold meshes with tufted cover and intrinsic Delaunay triangulations defined on a local neighborhood of each vertex~\cite{Sharp2019}.
However, applying the cotangent Laplacian operator to point clouds presents challenges due to the absence of connectivity information.

\paragraph{Laplacian Operator on Point Clouds}
The basic idea of defining a Laplacian operator on a point cloud in prior works is to first construct a local approximation of the underlying manifold.
This is essential as the Laplacian operator is a differential operator defined within a local manifold.
A prevalent approach is to build a local Delaunay triangulation of the point cloud using the $K$-nearest neighbors graph and subsequently build the Laplacian operator with the heat kernel on the triangulation~\cite{Belkin2009,Liu2012,Luo2009}.
The accuracy and convergence of the Laplacian operator on sharp features are further improved by anisotropic Voronoi diagrams~\cite{Qin2018}.
Some other methods only utilize local triangulations to determine point cloud connectivity and use the positions of 3D point clouds to compute the Laplacian weights~\cite{Clarenz2004,Cao2010}.
Alternatively, other methods leverage the smoothed particle hydrodynamics and global optimizations~\cite{Petronetto2013} or moving least squares~\cite{Liang2012} to approximate the underlying surface.
The method of~\cite{Sharp2020a} defines Laplacian operators on nonmanifold meshes, suggesting its applicability to point clouds using local triangulations.
However, these methods may face difficulties particularly when addressing noisy point clouds, thin structures, and sharp features.
In contrast, our method directly learns the Laplacian operator by understanding shapes, demonstrating significantly improved robustness and accuracy.

\paragraph{Graph Neural Networks}
Graphs serve as versatile data structures in geometry processing: meshes are essentially graphs, and point clouds can be readily converted to graphs by connecting the $K$-nearest neighbors.
Therefore, graph neural networks (GNNs) find extensive applications in numerous 3D contexts.
Representative works in mesh applications include employing GNNs for mesh understanding~\cite{Hanocka2019,Hu2022}, simulation~\cite{Pfaff2020}, subdivision~\cite{Liu2020c}, generation~\cite{Groueix2018}, and geodesic distance computation~\cite{Pang2023}.
Similarly, plenty of works also apply GNNs to point clouds for point cloud understanding~\cite{Qi2017,Li2018,Wang2019c}, generation~\cite{Fan2017,Groueix2018}, and normal estimation~\cite{Atzmon2018}.
In this paper, we utilize graph neural networks to learn the Laplacian operator on point clouds with the KNN graph.
The pivotal component of the GNN is the graph convolutional layer.
While various graph convolution designs have been proposed, such as the graph convolution layer~\cite{Kipf2017}, graph attention layer~\cite{Velickovic2017}, edge convolution~\cite{Gilmer2017}, and GraphSAGE~\cite{Hamilton2017}, we design a simple yet effective graph convolution layer tailored for our task, which has proven to be more effective than prevailing graph convolution layers.

\paragraph{Neural Operators}
Recently, neural networks have been used in learning operators on geometric shapes. Such operators are often designed to generalize traditional numerical methods for solving partial differential equations (PDEs).
Representative works include \cite{Wang2019d,Lu2019,Li2020a}, in which a nonlinear operators is learned from data and then used for downstream applications.
Some other works focus on efficiency \cite{Rahman2023} or large-scale scenes \cite{Li2024}.

%% file: src/method.tex
\section{Method} \label{sec:method}

\begin{figure*}[t]
    \centering
    \includegraphics[width=0.98\linewidth]{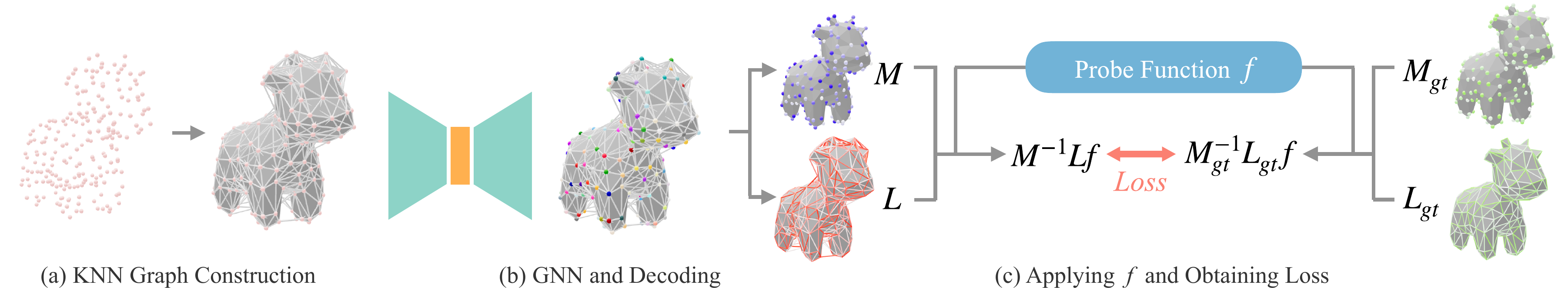}
    \caption{Overview. (a): a $K$-nearest neighbors graph (right) is constructed from the input unoriented point cloud (left).
    The underlying mesh is drawn together with $K$-nearest neighbors graph for better visualization.
    (b): a GNN is used to learn the vertex-wise features, followed by two MLPs to decode the features to a Laplacian matrix $L$ and a mass matrix $M$.
    (c): we compute the Laplacians of probe functions with the predicted $L$ and $M$ and the ground-truth $L_{gt}$ and $M_{gt}$ and minimize the difference between the two Laplacians to train the GNN.
    Although the connectivity of the KNN graph is different from the ground truth mesh, we can still compare the Laplacians of the probe functions to train the GNN.
    }
    \label{fig:overview}
  \end{figure*}

\paragraph{Overview}
We utilize a Graph Neural Network (GNN) to learn the Laplacian operator for an input point cloud.
An overview of our method is presented in \cref{fig:overview}.
Initially, we construct a KNN graph on the point cloud.
Instead of trying to improve the quality of graph topology, we employ a GNN to learn the Laplacian operator from data by mimicking the behavior of the ground-truth Laplacian operator.
In \cref{sec:laplacian}, we first introduce the formulation and the overall idea of our neural Laplacian operator.
Next, we detail the probe functions for evaluating and comparing the behavior of our Laplacian operator and the ground-truth Laplacian operator in \cref{sec:probe}.
Lastly, we introduce the details of the GNN in \cref{sec:gnn}.

\subsection{Neural Laplacian Operator} \label{sec:laplacian}

\paragraph{Discrete Laplacian Operator}
We formulate the definition of the discrete Laplacian operator within the context of graphs.
Polygonal meshes are essentially graphs, with each vertex connected to its neighboring vertices by edges.
Similarly, point clouds can be modeled as graphs, where each point serves as a vertex and is connected to its $K$-nearest neighbors by edges.
Consider a graph $\mathcal{G} = (\mathcal{V}, \mathcal{E})$, where $\mathcal{V}$ is the set of vertices, and $\mathcal{E}$ is the set of edges.
Each vertex in $\mathcal{V}$ is represented by $v_i$ with an index $i$, and each edge in $\mathcal{E}$ is represented by $(i,j)$, where $v_i, v_j \in \mathcal{V}$.
As a linear differential operator, the discrete Laplacian operator for a graph $\mathcal{G}$ can be depicted as a matrix.
Thus, we use the term Laplacian matrix and Laplacian operator interchangeably in this paper.
Generally speaking, the Laplacian matrix can be considered as the product of the inverse of the mass matrix and the stiffness matrix.
In the stiffness matrix, each element $L_{ij}$ corresponds to the weight of the edge $(i,j)$ and is defined as:
\begin{equation}
    L_{ij} =
    \begin{cases}
        \sum_{k=1}^{n} w_{ij},
            & \text{if } i = j    \text{ and } (i,j) \in \mathcal{E}; \\
        - w_{ij},
            & \text{if } i \neq j \text{ and } (i,j) \in \mathcal{E}; \\
        0,
            & \text{otherwise}.
    \end{cases}
    \label{eq:laplacian}
\end{equation}
Accompanying the stiffness matrix $L$ is the mass matrix $M$, which is a diagonal matrix with each diagonal element $M_{ii}$ as the mass of vertex $v_i$.

\begin{wrapfigure}[7]{r}{0.26\linewidth}
    \centering
    \vspace{-6pt}
    \hspace{-22pt}
    \includegraphics[trim=0 0 0 0,clip,width=1.1\linewidth]{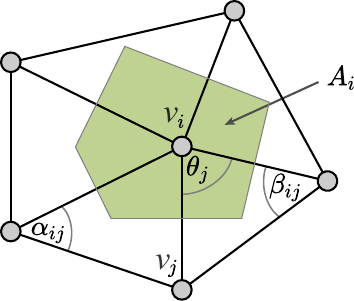}
\end{wrapfigure}
For general graphs, the edge weight $w_{ij}$ ($i \neq j$) and $M_{ii}$ can be set to 1, yielding the unweighted Laplacian operator~\cite{Taubin1995}.
In the case of triangular meshes, the edge weight $w_{ij}$ can be determined as the cotangent of the angle between the two adjacent edges~\cite{Pinkall1993,Desbrun1999,Meyer2003}, and the mass $M_{ii}$ can be assigned the value corresponding to the area of the Voronoi cell incident to vertex $v_i$.
This configuration defines the well-known cotangent Laplacian operator:
\begin{equation}
    w_{ij} = \cot(\alpha_{ij}) + \cot(\beta_{ij}), \quad M_{ii} = A_i,
    \label{eq:cot-laplacian}
\end{equation}
where the definition of $\alpha_{ij}$, $\beta_{ij}$, and $A_i$ are shown in the inset figure.
For point clouds, prevalent techniques first construct a local triangulation with the $K$-nearest neighbors around each point, then create the Laplacian operator using the triangulation, adhering to principles akin to the discretization of the Laplacian operator on triangular meshes~\cite{Belkin2009,Luo2009,Liu2012,Sharp2020a}.

\paragraph{Neural Laplacian Operator}
We aim to extend the discrete cotangent Laplacian to point clouds with high accuracy.
For an input point cloud, we assume that it is sampled from an underlying manifold triangle mesh.
We first construct a KNN graph on the point cloud, connecting each point to its $K$-nearest neighbors.
Then, we augment the KNN graph by adding edge $(j,i)$ to the graph if $(i,j)$ is an edge to ensure symmetry, and we also add a self-loop to each vertex to ensure that the diagonal elements of the Laplacian matrix are non-zero.
We set $K$ to 8 for KNN graph construction to ensure the sparsity of $L$.
Our goal is to employ a GNN to learn the edge weights $w_{ij}$ and the mass $M_{ii}$ for point clouds on the KNN graph, such that the resulting Laplacian matrix $L$ and mass matrix $M$ have similar behavior to the ground-truth Laplacian matrix $L_{gt}$ and mass matrix $M_{gt}$ defined on the corresponding manifold mesh.
Since the connectivity of the KNN graph and the ground-truth mesh are different, we cannot enforce $L$ to be identical to $L_{gt}$.
Instead, we encourage them to exhibit similar behaviors on a set of functions. Specifically, for an arbitrary function $f$ defined on the mesh, we can compute its Laplacian as:
\begin{equation}
    \Delta_{gt} f = M^{-1}_{gt} L_{gt} f.
\end{equation}
As we construct the point cloud dataset using the vertices of the mesh, mapping the function $f$ to the point cloud becomes a trivial task.
Then, we can also compute $\Delta f$ on the point cloud with the learned $L$ and $M$.

We train the neural Laplacian operator by minimizing the difference between $\Delta f$ and $\Delta_{gt} f$:
\begin{equation}
    \mathcal{L}_{laplacian} = \sum_{f \in \mathcal{F}}
              w_f \| M^{-1} L f - M^{-1}_{gt} L_{gt} f \|^2_2,
    \label{eq:loss}
\end{equation}
where $\mathcal{F}$ is the function space defined on the 3D shape, and $w_f$ is a weight to balance contribution of a function $f$ to the loss:
\begin{equation}
    w_f = \frac{1}{mean\left(M_{gt}^{-1} L_{gt} f\right) + \epsilon},
\end{equation}
where $\epsilon$ is set to $0.1$ to avoid numerical instability.
We name the functions in $\mathcal{F}$ as \emph{probe functions} since they are used to probe the behavior of the Laplacian operator.
In addition, we empirically find that adding an additional loss on the diagonal of the mass matrix $M$ can make the training converge slightly faster:
\begin{equation}
    \mathcal{L}_{mass} = mean \left( \| diag(M) - diag(M_{gt}) \|^2 \right),
\end{equation}
Therefore, the overall loss is:
\begin{equation}
    \mathcal{L} = \mathcal{L}_{laplacian} + \lambda \mathcal{L}_{mass},
\end{equation}
where $\lambda$ is a hyperparameter to balance the two terms and is set to 0.1 in our experiments.

The GNN takes the KNN graph as input and outputs a feature vector $p_i$ for each vertex $v_i$, as shown in \cref{fig:overview}.
Then we use two MLPs to predict the edge weight $w_{ij}$ and the mass $M_{ii}$ from point features: \looseness=-1
\begin{align}
    w_{ij} &= \Phi_{edge}\left(\left( p_i - p_j \right)^2 \right), \label{eq:edge} \\
    M_{ii} &= \Phi_{mass}\left( p_i \right), \label{eq:mass}
\end{align}
where $\Phi_{edge}$ and $\Phi_{mass}$ are two lightweight MLPs with Softplus activation.
\rev{Note that $\Phi_{edge}$ takes the square of the element-wise difference of $p_i$ and $p_j$ to ensure the symmetry of $w_{ij}$ concerning $v_i$ and $v_j$.}
We also append a ReLU and a Softplus activation function to the output of $\Phi_{edge}$ and $\Phi_{mass}$ to ensure the edge weight and the mass are positive.

\subsection{Probe Functions} \label{sec:probe}

The probe functions $\mathcal{F}$ are the key to the training of the GNN.
On one hand, the chosen probe functions must exhibit sufficient diversity to encompass all functions defined on a 3D shape, which ensures that our neural Laplacian operator can be applied to arbitrary functions.
On the other hand, these probe functions should be independent of each other to prevent redundancy and enhance training efficiency.
To achieve this objective, we devise two groups of probe functions for each 3D shape, including \emph{spectral} and \emph{spatial} probe functions.

\begin{figure}[t]
    \centering
    \includegraphics[width=0.91\linewidth]{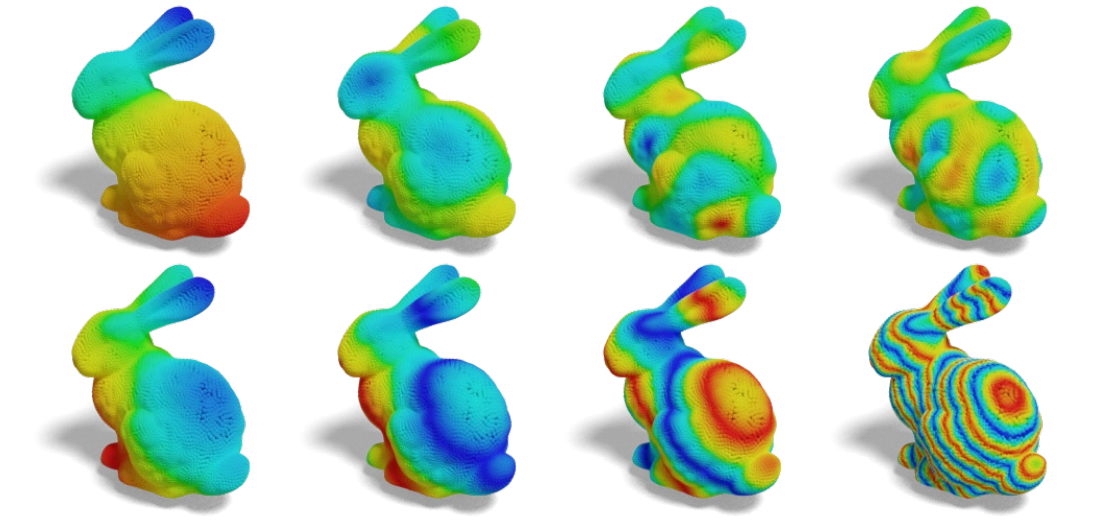}
    \caption{Visualization of probe functions.
    The top row shows spectral probe functions, consisting of the 1st, 16th, 32th, and 64th non-constant eigenvectors of the ground-truth Laplacian matrix.
    The bottom row shows sample spatial probe functions with different frequencies and phases.
    }
    \label{fig:probes}
\end{figure}

\paragraph{Spectral Probe Functions}
Theoretically, the eigenfunctions of the ground-truth Laplacian operator are linearly independent and span the entire function space on a 3D shape.
However, it is neither practical nor necessary to utilize all the eigenfunctions as probe functions.
Firstly, solving all eigenfunctions of a Laplacian operator is implemented via the eigen decomposition of the Laplacian matrix, which is computationally expensive, especially for large meshes with tens of thousands of points.
Secondly, the eigenfunctions with large eigenvalues are of high frequency and prone to being adversely affected by noise. Thus, they are not advantageous for training the network.
Also, previous work~\cite{Levy2006} suggests that the shape information is typically encoded in the eigenfunctions with small eigenvalues.
Therefore, in our experiments, we employ the first 64 eigenfunctions of the ground-truth Laplacian operator as spectral probe functions, excluding the eigenvector with a zero eigenvalue.
Empirically, we find that this approach is sufficient to achieve good performance.
In addition, to prevent the training of the network from being dominated by eigenfunctions with large eigenvalues, we scale each eigenfunction with its eigenvalues: \looseness=-1
\begin{equation}
    f_i = \frac{f_i}{\lambda_i + \epsilon_1},
\end{equation}
where $f_i$ is the $i$-th spectral probe function, $\lambda_i$ is the corresponding eigenvalue, and $\epsilon_1$ is set to 0.1 to avoid numerical instability.
We name this type of functions as spectral probe functions since they are the eigenfunctions of the Laplacian operator.
The visualization of several spectral probe functions is shown in the first row of \cref{fig:probes}.

\paragraph{Spatial Probe Functions}

To capture the high-frequency behaviors of the Laplacian operator, we introduce a set of spatial probe functions.
Specifically, to compensate for the lack of high-frequency behavior in the spectral probe functions, we train the network using the following function $f$ as spatial probe functions:
\begin{equation}
f(p) = \frac{1}{2 k} \sin \left( k \psi \left( ax+by+cz \right) + \phi \right),
\end{equation}
where $k \in \{ 2^{m/2} | m=0,1,2,3...13 \}$, $\psi$ represents frequency noise uniformly sampled in $[0.75, 1.25]$, $\phi$ is a random phase in $[0, 2\pi]$, and $a, b, c$ are randomly sampled such that their sum is 1. The term $\frac{1}{2 k}$ is a normalization factor to ensure that the spatial probe functions will not be dominated by $f$ with large $k$.
The visualization of several spatial probe functions is presented in the second row of \cref{fig:probes}.
Note that we ignore the zero eigenvalue and the corresponding all-constant eigenvector in our experiments since this eigenvector is all $\mathbf{0}$, thus not informative for the training of the Laplacian operator.

\paragraph{Combination of Probe Functions}
During training, we concatenate the spectral and spatial probe functions, apply our learned Laplacian to them, and minimize the loss as defined in \cref{eq:loss}.
Theoretically, all discrete functions defined on a point cloud can be expanded as a linear combination of the eigenvectors of the Laplacian matrix.
Thus, we use eigenvectors so that the behaviors of NeLo and the ground-truth Laplacian can be as similar as possible under a wide range of functions.
However, computing all eigenvectors is expensive; we use several low-frequency eigenvectors and compensate for the effect of high-frequency eigenvectors with high-frequency sinusoids.
The experiments demonstrate that the combination of the two types of probe functions yields better performance compared to using either one alone.
This synergy enables our neural Laplacian operator to effectively operate on a diverse range of functions defined on the point cloud, which enhances the feasibility of applying Laplacian-based geometry processing algorithms to point clouds.

\subsection{Graph Neural Network} \label{sec:gnn}

\begin{figure}[t]
    \centering
    \includegraphics[width=0.99\linewidth]{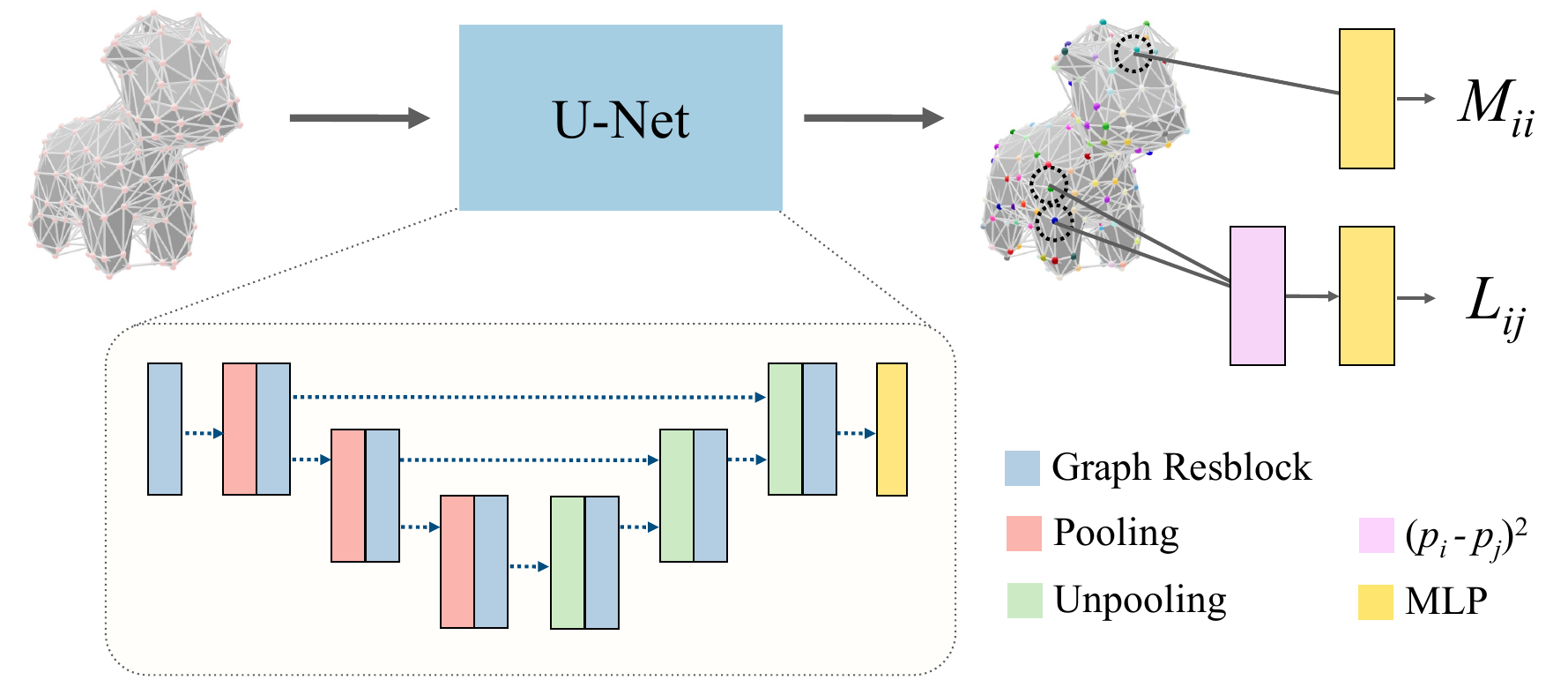}
    \caption{The network architecture.
    The U-Net takes a KNN graph as input and outputs a feature vector for each vertex.
    Then, two MLPs are used to predict the edge weights of the stiffness matrix and elements of the mass matrix, respectively.
    }
    \label{fig:network}
\end{figure}

We employ a U-Net~\cite{Ronneberger2015} for learning on the KNN graph, and the overall network architecture is shown in \cref{fig:network}.
In this section, we provide a detailed description of the key components of the network.
We follow classical GNN approaches, which naturally cope with varying numbers of input points.\looseness=-1

\paragraph{Input Signal}\label{sec:method:input_signal}
A good Laplacian operator should be translation-invariant and scalable. Previous geometric deep learning works usually take the absolute coordinates of each vertex $(x, y, z)$ as input signal \cite{Qi2017,Wang2019c}.
However, the Laplacian operator is a local operator by nature, and using global coordinates will undermine the translation-invariance and scalability of the network from the very foundation.
To address this issue, we propose to simply feed the network with an input signal of all-one vector $(1,1,1)$, \rev{similar to \cite{Li2020b}.}
We additionally concat the number of neighbors of each vertex to the input signal, which is trivially obtained from the KNN graph.
This approach deliberately withholds global geometric information from the network. This compels the network to learn the translation-invariant, local information during the message-passing process.

\paragraph{Graph Convolution}\label{graph_convolution_operator_design}
Our graph convolution is designed under the message-passing paradigm on the graph~\cite{Gilmer2017,Simonovsky2017,Fey2019}.
Inspired by ~\cite{Pang2023,Hamilton2017,Sanchez2020}, we design a graph convolution operator that takes the geometry properties of edges into consideration.
Specifically, it takes relative positions between vertices and edge length as additional features when performing convolution, which is helpful for the network to understand local geometry structures.
It proves to be more effective for our task than other popular designs.
Specifically, for each vertex $v_i$, we aggregate the features of its neighbors $v_j \in \mathcal{N}(i)$ and itself to update the feature of $v_i$, while considering the relative position and distance between $v_i$ and $v_j$:
\begin{equation}
    p_i = W_0 \  p_i +
        \sum \nolimits_{j \in \mathcal{N}(i)}
        \left( W_1 \cdot [p_j \mathbin\Vert v_{ij} \mathbin\Vert l_{ij}] \right),
    \label{eq:conv}
\end{equation}
where $p_i$ and $p_j$ are the features of $v_i$ and $v_j$, $W_0$ and $W_1$ are learnable weights, $\mathbin\Vert$ is a concatenation operator, $v_{ij}$ denotes $v_i - v_j$, and $l_{ij}$ is the length of $v_{ij}$.

\paragraph{Graph Downsampling and Upsampling}
In addition to graph convolution, graph downsampling, and upsampling are indispensable for the U-Net.
For downsampling, we utilize voxel-based pooling~\cite{Simonovsky2017}, which overlays a voxel grid on the point cloud and computes the average of vertices and features in each voxel as that of the downsampled graph.
Upsampling is accomplished through interpolation by reversing the voxel-based pooling mapping.
The side length of a voxel for downsampling is set to $\frac{1}{16}$ at the first layer and doubles at each subsequent downsampling layer.

\paragraph{Detailed Configurations}
We first construct residual blocks (resblocks)~\cite{He2016} with a sequence of graph convolutions, Group Normalizations~\cite{Wu2018d}, and ReLU activations.
Then, we stack residual blocks, downsampling, and upsampling blocks to form the U-Net.
The detailed configurations of the network are shown in \cref{fig:network}.
For the residual blocks, ranging from fine to coarse levels, the numbers of input feature channels are 128, 128, 128,  the output feature channels are 256, 256, 512, and the numbers of blocks are 3, 2, 3.
Note that the layers and channels are hyperparameters and could be tuned for better performance.
The U-Net takes all one vectors as input and generates a 256-feature vector $p_i$ for each vertex $v_i$ as output.
The MLPs predicting edge weights $\Phi_{edge}$ and mass $\Phi_{mass}$ consist of two fully connected layers with 256 hidden units and a Softplus activation function in between.

%% file: src/results.tex
\section{Results} \label{sec:result}

In this section, we compare our method with previous approaches, demonstrating its effectiveness in \cref{sec:experiments}.
We also apply our method on real-world point clouds in \cref{sec:real_world_scenes}.
We delve into the effects of key components of our method and discuss the properties of our neural Laplacian operator in \cref{sec:ablation}.
And we apply it to a series of geometry processing tasks in \cref{sec:application}.

\subsection{Evaluations and Comparisons} \label{sec:experiments}

\paragraph{Dataset}
We utilize a subset of ShapeNet as our dataset, containing $12k$ 3D shapes across 17 categories.
We randomly split them into training and testing sets with an approximate ratio of $80\%:20\%$.
For data preprocessing, we first convert the meshes into watertight manifold meshes using the method in~\cite{Wang2022}.
\rev{For meshes with open boundary, please refer to \cref{sec:real_world_scenes}.}
Subsequently, we employ mesh simplification~\cite{Garland1997} and remeshing~\cite{Hoppe1993} to get the final meshes with an average vertex number of $5.2k$.
We normalize the meshes to the range of $[-1, 1]$ and discard the connectivity to get the corresponding point cloud for training.
We use the libigl library~\cite{Libigl2018} to compute the cotangent Laplacian and mass matrix on the mesh following~\cref{eq:cot-laplacian}, which serve as the ground truth.
Since the diagonal values of the mass matrix are numerically small and vary in different data, we normalize them by dividing them by their mean.

\begin{table*}[t]
    \tablestyle{4pt}{1.2}
    \caption{Quantitative comparisons on ShapeNet.
    The MSE, the rate of MSE larger than 1.0 ($R_{MSE>1}$), and the average number of non-zero elements per row in the resulting Laplacian matrices (Sparsity) are reported for each method.
    We list the detailed statistics of some large categories in our dataset, as well as the results over all 17 categories (\textbf{Total}). We further report the improvement ratio of our method over the best baseline in terms of MSE and $R_{MSE>1}$ in \hl{green} color.
    Our method reduces the MSE by an order of magnitude compared with existing methods, and significantly outperforms them in terms of $R_{MSE>1}$.
    }
    \begin{tabular}{llcccccccccccccccc}
        \toprule
    \multirow{2}{*}{Category} & \multirow{2}{*}{\#Num.} & \multicolumn{3}{c}{Graph\scriptsize{~\cite{Taubin1995}}} && \multicolumn{3}{c}{Heat\scriptsize{~\cite{Belkin2009}}} && \multicolumn{3}{c}{NManifold\scriptsize{~\cite{Sharp2020a}}} && \multicolumn{3}{c}{Ours} \\
    \cline{3-5}\cline{7-9}\cline{11-13}\cline{15-17}
     &  & MSE & $R_{MSE>1}$ & Sparsity& & MSE & $R_{MSE>1}$ & Sparsity && MSE & $R_{MSE>1}$ & Sparsity& & MSE & $R_{MSE>1}$ & Sparsity  \\ \midrule

     Airplane & 272 & 0.4132 & 19.70\% & \hspace{-0.5em}10.0 & ~ & 0.4191 & 24.36\% & 11.0 & ~ & 0.1154 & 3.06\% & 8.4 & ~ & 0.0042 & 0.02\% & \hspace{-0.5em}10.0 \\
        Bag & 10 & 0.1779 & \hspace{0.5em}6.70\% & 9.8 & ~ & 0.1897 & \hspace{0.5em}9.55\% & 10.2 & ~ & 0.0669 & 3.93\% & 7.8 & ~ & 0.0045 & 0.00\% & 9.8 \\
        Bed & 23 & 0.2599 & 11.76\% & 9.9 & ~ & 0.3757 & 25.74\% & 10.5 & ~ & 0.1389 & 7.14\% & 8.4 & ~ & 0.0106 & 0.39\% & 9.9 \\
        Bench & 207 & 0.2985 & 17.53\% & 9.8 & ~ & 0.3108 & 22.82\% & 10.3 & ~ & 0.2149 & \hspace{-0.5em}11.69\% & 9.1 & ~ & 0.0121 & 0.38\% & 9.8 \\
        Bookshelf & 59 & 0.3092 & 17.33\% & 9.8 & ~ & 0.3008 & 23.05\% & 10.0 & ~ & 0.1705 & 5.08\% & 9.3 & ~ & 0.0030 & 0.05\% & 9.8 \\
        Bottle & 51 & 0.1602 & \hspace{0.5em}4.67\% & 9.7 & ~ & 0.1040 & \hspace{0.5em}7.32\% & \hspace{0.5em}9.3 & ~ & 0.0663 & 5.04\% & 7.2 & ~ & 0.0028 & 0.11\% & 9.7 \\
        Cabinet & 210 & 0.1581 & \hspace{0.5em}6.70\% & 9.8 & ~ & 0.1511 & 10.70\% & \hspace{0.5em}9.5 & ~ & 0.0718 & 3.25\% & 7.7 & ~ & 0.0011 & 0.01\% & 9.8 \\
        Phone & 8 & 0.0887 & \hspace{0.5em}0.00\% & 9.7 & ~ & 0.0144 & \hspace{0.5em}0.00\% & \hspace{0.5em}9.1 & ~ & 0.0259 & 0.22\% & 7.6 & ~ & 0.0008 & 0.00\% & 9.7 \\
        Chair & 172 & 0.2100 & 10.04\% & 9.8 & ~ & 0.2028 & 13.22\% & 10.1 & ~ & 0.1243 & 6.84\% & 8.3 & ~ & 0.0058 & 0.19\% & 9.8 \\
        Display & 175 & 0.1776 & \hspace{0.5em}6.65\% & 9.8 & ~ & 0.1768 & 10.94\% & \hspace{0.5em}9.6 & ~ & 0.1074 & 5.82\% & 8.1 & ~ & 0.0036 & 0.08\% & 9.8 \\
        Lamp & 190 & 0.3606 & 19.54\% & 9.8 & ~ & 0.3471 & 25.05\% & 10.6 & ~ & 0.1760 & 9.59\% & 8.5 & ~ & 0.0071 & 0.12\% & 9.8 \\
        Mailbox & 10 & 0.1902 & \hspace{0.5em}3.13\% & 9.8 & ~ & 0.1783 & 10.54\% & \hspace{0.5em}9.5 & ~ & 0.0252 & 0.36\% & 7.4 & ~ & 0.0019 & 0.00\% & 9.8 \\
        Sofa & 222 & 0.1513 & \hspace{0.5em}6.08\% & 9.8 & ~ & 0.1461 & \hspace{0.5em}9.64\% & \hspace{0.5em}9.7 & ~ & 0.0728 & 4.26\% & 7.5 & ~ & 0.0030 & 0.05\% & 9.8 \\
        Table & 230 & 0.2124 & 10.30\% & 9.8 & ~ & 0.2092 & 14.36\% & 10.0 & ~ & 0.1474 & 7.19\% & 8.8 & ~ & 0.0042 & 0.10\% & 9.8 \\
        Telephone & 232 & 0.1157 & \hspace{0.5em}2.19\% & 9.7 & ~ & 0.0538 & \hspace{0.5em}3.14\% & \hspace{0.5em}9.1 & ~ & 0.0408 & 2.54\% & 7.4 & ~ & 0.0011 & 0.03\% & 9.7 \\
        Train & 47 & 0.2689 & \hspace{0.5em}7.45\% & 9.9 & ~ & 0.2749 & 13.89\% & 10.5 & ~ & 0.1347 & 7.07\% & 7.9 & ~ & 0.0070 & 0.15\% & 9.9 \\
        Watercraft & 284 & 0.3003 & 12.77\% & 9.9 & ~ & 0.3162 & 20.36\% & 10.5 & ~ & 0.1248 & 5.92\% & 8.2 & ~ & 0.0066 & 0.19\% & 9.9 \\

    \midrule

        \textbf{Total}. & 2402 & 0.2444 & 11.17\% & 9.8 & ~ & 0.2386 & 15.68\% & 10.1 & ~ & 0.1181 & 5.80\% & 8.2 & ~ & 0.0049 \hl{($\times 24$)}  & 0.12\% \hl{($\times 50$)} & 9.8 \\

    \bottomrule
    \end{tabular}
    \label{tab:shapenet_comparison}
\end{table*}

\paragraph{Training Details and \rev{Inference Time}}\label{training_datails}
We utilize the AdamW optimizer~\cite{Loshchilov2017} for training with an initial learning rate of 0.001 and a weight decay of 0.01, which are the default settings in PyTorch.
We gradually reduce the learning rate to 0 throughout the training process with a linear decay schedule.
In our implementation, the network is trained with a batch size of 8 for 500 epochs, which takes approximately 16 hours on 4 Nvidia 4090 GPUs with 24GB of memory.
After training, we only need a one-time feed-forwarding \rev{inference} to get the Laplacian matrix of a point cloud. \rev{On a single Nvidia 4090 GPU, for most point clouds in our dataset, the time consumption of such feed-forwarding is around 90ms and takes about 1G GPU memory at peak.
The time of constructing a KNN Graph is around 100 ms on CPU.}

\begin{figure*}[t]
    \centering
    \includegraphics[width=0.94\linewidth]{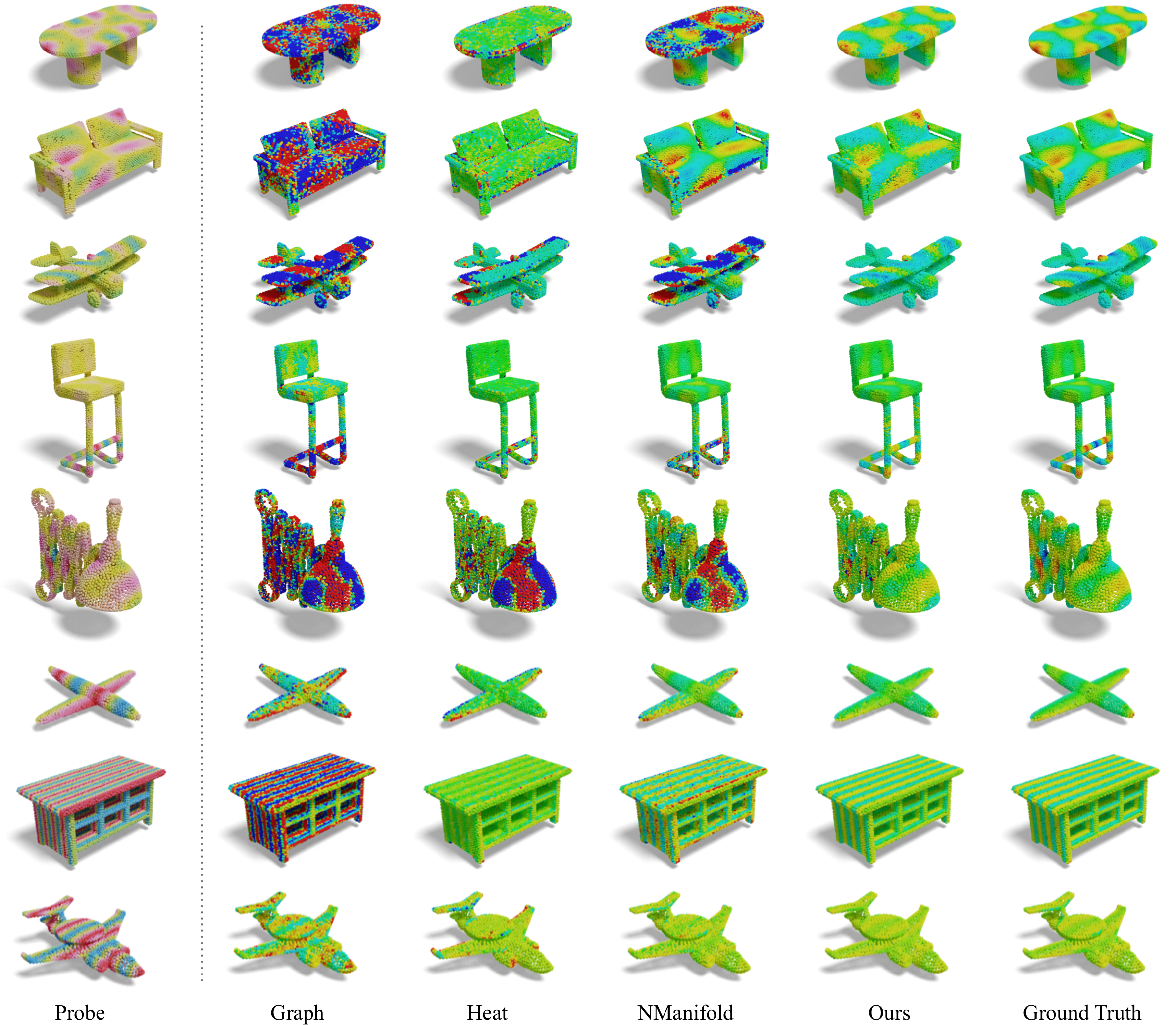}
    \caption{Visual comparisons.
    From left to right are the probe functions, results of \emph{Graph}~\cite{Taubin1995}, \emph{Heat}~\cite{Belkin2009}, \emph{NManifold}~\cite{Sharp2020a}, our method, and the ground truth Laplacian operator, respectively.
    Our results are apparently more faithful to the ground truth, especially for the point clouds with sharp features and thin structures.
    }
    \label{fig:comparison}
\end{figure*}

\begin{figure*}[t]
    \centering
    \includegraphics[width=0.95\linewidth]{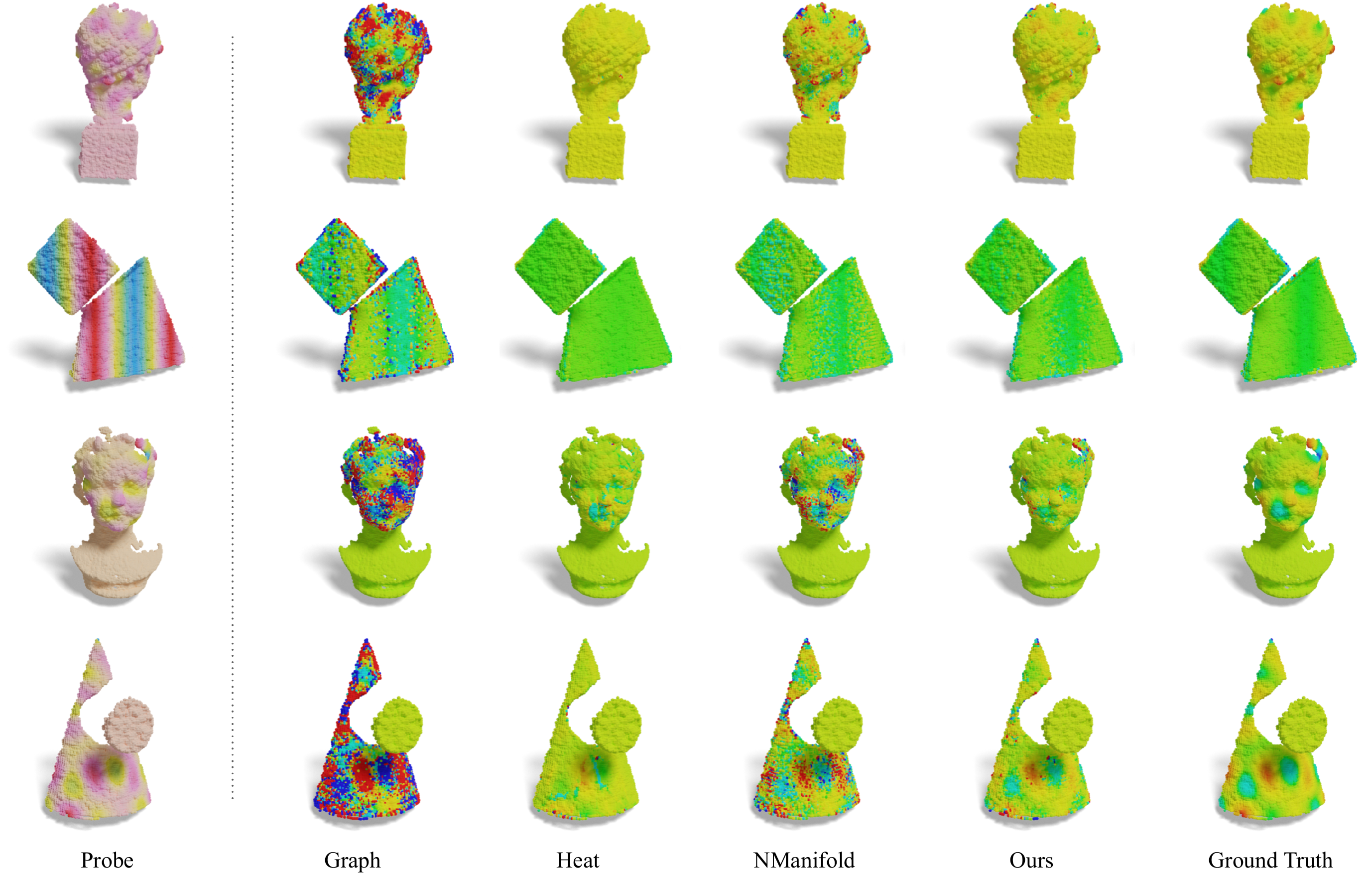}
    \caption{The visualization of the results on real-world data.
    The point cloud is collected in real scenes by a Microsoft Kinect v2 \cite{Wang2016a}. From left to right are the probe functions, results of \emph{Graph}~\cite{Taubin1995}, \emph{NManifold}~\cite{Sharp2020a}, our method, and the ground truth Laplacian operator, respectively.
    As the image suggests, our method is robust for wild scenes as well.
    }
    \label{fig:real_world_scenes}
\end{figure*}

\begin{figure}[h]
    \centering
    \includegraphics[width=0.97\linewidth]{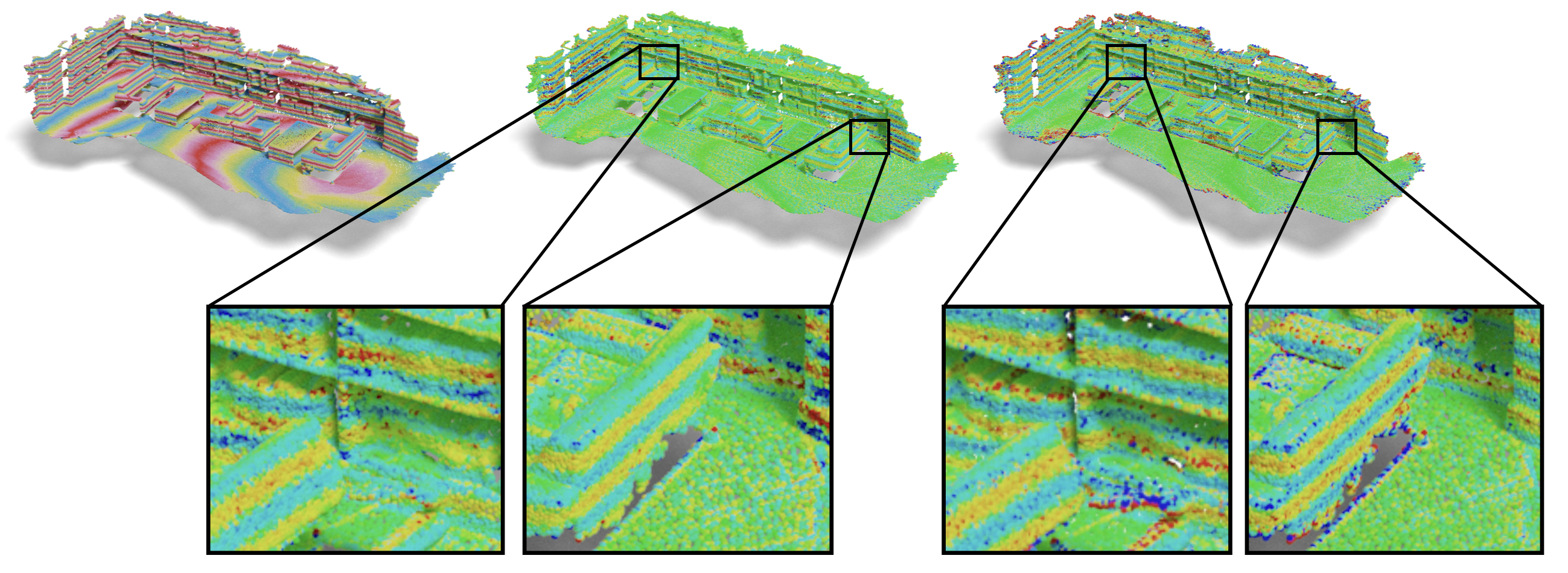}
    \caption{Generalization and Scalability.
    To show the generalization ability of our method, we directly apply our method to a ScanNet scene, without any fine-tuning or retraining. For the given probe function (left), we visualize our result (middle) and a reference result (right), \rev{of which the Laplacian is obtained by applying \cite{Sharp2020a} on the reference mesh}.
    }
    \label{fig:generalization_and_scalability}
\end{figure}

\paragraph{Evaluation Metrics}\label{evaluation_metrics}
We evaluate a Laplacian operator by applying it to a probe function and calculating the Mean Squared Error (MSE) between the result and the result of the ground-truth Laplacian operator.
\rev{Specifically, for each sample, we apply \cref{eq:loss} with its balance weight removed to calculate the per-point MSE loss, then we take the average of all points as the MSE of the mesh.}
For each evaluated Laplacian operator, we select 112 probe functions, including the first 64 non-constant eigenvectors of the ground-truth Laplacian matrix and 48 spatial probe functions.
The 42 of spatial probe functions are defined as $f(t)=\frac{1}{2k}sin(kt + \phi)$, where $t$ is from $\{x, y, z\}$, $k$ is from $\{1,2,4,8,16,32,64\}$, and $\phi$ is from $\{0, \pi/2\}$.
The remaining spatial probe functions are polynomials: $x, y, z, x^2, y^2, \text{and}\; z^2$.
Notably, polynomial probe functions are unseen during training and thus could be an indicator of the generalization ability in terms of different probe functions.
Then, we compute MSEs on point clouds from the testing set, comprising 2402 models from 17 categories in ShapeNet.
We observe that the mean squared error (MSE) can be inflated by outliers in certain test cases. To mitigate this impact and achieve a more fair evaluation, we employ an MSE clipping.
For each sample and probe function combination, we apply a clipping threshold of 1.0, which is approximately 20 times the average MSE observed across our method's predictions.
We also report the percentage of samples with MSEs exceeding 1.0, denoted as $R_{MSE>1}$, which provides insight into the robustness of the evaluated Laplacian operator.
Additionally, the sparsity of the Laplacian matrix is an important factor in the efficiency of the downstream algorithms. Thus, we report the sparsity of the resulting Laplacian matrices, which is defined as the average number of non-zero elements per row in the resulting Laplacian matrices.
For visual comparisons, we keep the color range of the images from different algorithms the same, with the maximum and minimum values of the color range being the maximum and minimum values of the results of the ground truth Laplacian operator.  \looseness=-1

\paragraph{Comparisons}
We evaluate the performance of our neural Laplacian operator against previous methods using the evaluation metrics above. The comparison includes:
\begin{itemize}[leftmargin=16pt,itemsep=2pt]
    \item[-] \emph{Graph}: An uniform weight KNN graph constructed from the point cloud. It is widely used in graph and geometry applications~\cite{Taubin1995}. This is the simplest form of the Laplacian operator on a graph and can be considered as a naive baseline.
    \item[-] \emph{Heat}: The Laplacian operator proposed in~\cite{Belkin2009}, based on local triangulation and heat diffusion. \rev{We tuned its parameter such that it outputs a Laplacian matrix with basically the same sparsity as our method}.
    \item[-] \emph{NManifold}: The Laplacian operator in~\cite{Sharp2020a} which utilizes intrinsic Delaunay triangulation, currently representing the state-of-the-art method. We use the publicly available code provided by the authors. \rev{We use its default parameters since changing its hyperparameters usually has little effect. }
\end{itemize}
The quantitative results are summarized in \cref{tab:shapenet_comparison}, and the visual results are illustrated in \cref{fig:comparison}.
Our method consistently outperforms all other methods in terms of MSE and the rate of MSE larger than 1.0 ($R_{MSE>1}$).
Our method reduces the MSE by \textbf{\emph{an order of magnitude}} compared with existing methods.
Also, the rate of MSE larger than 1.0 is significantly lower than other methods.
The \emph{Graph} performs poorly since it ignores the geometric information of the point cloud.
The \emph{Heat} and \emph{NManifold} rely on local triangulations, making them challenging to work well on point clouds with sharp features or thin structures.
Representative examples include airplane wings, desk legs, flat tablets, and table corners, as shown in \cref{fig:comparison}.
In contrast, our method excels in handling these cases; our results are more faithful to the ground truth Laplacian operator.
The key reason is that our method learns geometric priors from data with a GNN and is not limited by local triangulation or graph connectivity.

\paragraph{Generalization}
Our neural Laplacian operator has a strong generalization ability to unseen shapes.
Although we train the network on a subset of ShapeNet, it performs well on meshes significantly different from ShapeNet.
ScanNet~\cite{Dai2017a} is a large-scale dataset of 3D reconstructions of indoor spaces created from 3D scans. It shares no overlap with ShapeNet in terms of shape categories and has far more vertices for each point cloud. To validate the generalization ability of our method, we \emph{directly} apply our pretrained model on a ScanNet scene, as in \cref{fig:generalization_and_scalability}.
The results are visually reasonable, demonstrating the good generalization ability of our method.

\paragraph{Scalability}
Our method is scalable to point clouds with varying sizes.
The scalability of NeLo originates from the fully convolutional nature of our network.
First, our network is a fully convolutional network, which is known for the ability to process inputs of varying sizes without the need for resizing or cropping.
Second, for every point in the input point cloud, we use an all-one vector as the input signal without any global coordinates information.
Such input signal, combined with the message-passing operator design in \cref{graph_convolution_operator_design}, makes our GNN focus on learning relative relations between points and local geometry, therefore, our network is capable of handling arbitrarily large input point cloud, providing that the system memory is sufficient.
As shown in \cref{fig:generalization_and_scalability}, we test the scalability of our method on a point cloud with 230k points, which is about $\times 50$ times larger than the averge number of points in training data.
We scale the point cloud so that the average edge length is approximately the same as that in the training data.
Our method performs well on the large-scale point cloud, though it has never been exposed to points of such pattern during training.

\begin{figure*}[ht]
    \centering
    \includegraphics[width=0.97\linewidth]{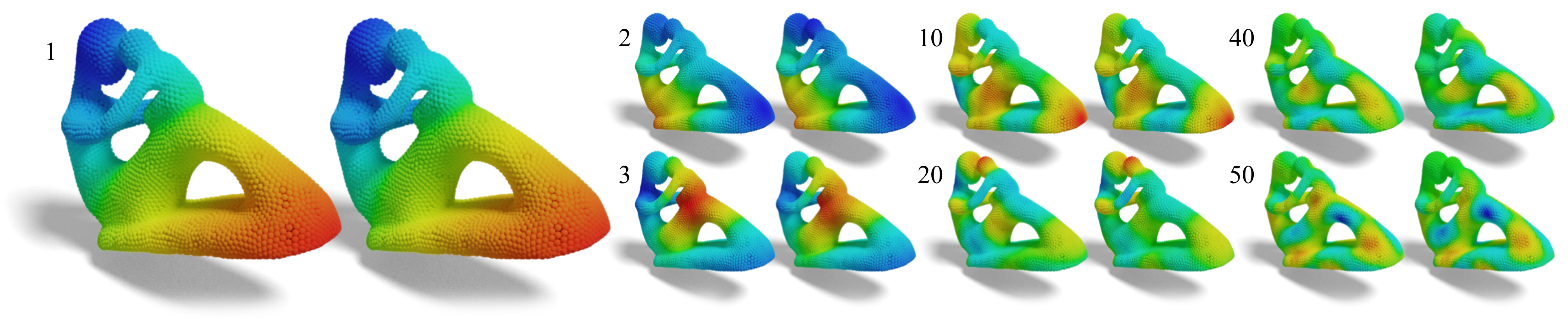}
    \caption{Eigen Decomposition of the neural Laplacian operator.
    Following~\cite{Nasikun2018}, each image pair depicts our prediction on the left and the ground truth on the right.
    The index of the non-constant eigenvector is displayed in the top left of each pair of image. }
    \label{fig:eigen_vecs}
\end{figure*}

\begin{figure}[t]
    \centering
    \includegraphics[width=0.95\linewidth]{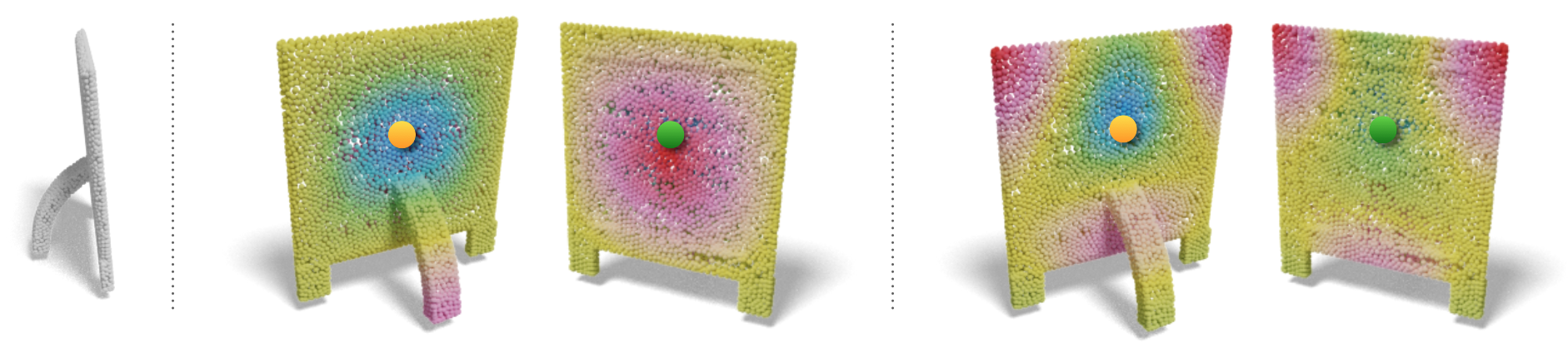}
    \caption{
    We visualizated the probe functions on an object with a large thin plate structure to show the importance of spectral probe functions in distinguishing the two opposite surfaces of thin structures.
    We rendered the 5th and 10th spectral probe functions, from front and back views, respectively.
    }
    \label{fig:spectral_probe_advantage}
\end{figure}

\paragraph{Eigen Decomposition of the Laplacian}
Our neural Laplacian operator exhibits excellent spectral properties.
The eigenvectors and eigenvalues closely match those of the ground truth Laplacian operator.
For the fertility model, we calculate the eigenfunctions of its predicted Laplacian, and compare it with the ground truth eigenfunctions.
Specifically, we first solve the generalized eigenvalue problem $Lx = \lambda Mx$, then we visualize the eigenvectors in \cref{fig:eigen_vecs}.
The results are almost visually indistinguishable from the ground truth Laplacian operator.
Note that the fertility model is not included in the training set, which demonstrates the good spectral property of our method.
Downstream geometry processing algorithm that relies on the eigenfunctions of the Laplacian matrix can benefit from such accurate spectral properties.

\subsection{Experiments on Real Scans} \label{sec:real_world_scenes}

In this section, we evaluate the performance of our method on the noisy real-world point clouds to demonstrate its robustness and practical utility.

\paragraph{Dataset and Preparation}
We use the dataset provided by~\cite{Wang2016a}, which is scannned using Microsoft Kinect v2.
This dataset contains a total of 144 scanned point clouds, already seperated into training and testing set, each with 50\% of all samples.
We followed their splitting settings for training and evaluation.
These point clouds are derived from raw depth images, exhibiting sparsity, varying sampling-densities, open-boundaries, noise, and holes, which puts a challenge for the Laplacian operator estimation.
For each noisy point cloud, the dataset also contains a corresponding clean mesh scanned by a high-precision professional Artec scanner, which is manually aligned with the noisy point cloud and can serve as a reference ground-truth model.
For training, we feed the scanned noisy point cloud to the network as input, and use the the Laplacian constructed from corresponding high-precision mesh as the ground truth.
Due to the limited number of samples in this dataset, we used the weights obtained in ~\cref{sec:experiments} as a starting point for the training.
Other experiment settings and evaluation metrics are exactly the same as in ~\cref{training_datails}.
We use a batch size of 2 and use 2 GPUs for training, which takes about 1 hour to converge.

\paragraph{Results}
We show the visual and quantitative comparisons in \cref{fig:real_world_scenes} and \cref{tab:real_world_scene_table}, respectively.
Our approach demonstrates significantly better performance than other methods when applied to real-world point clouds.
The real-world point clouds contain severe noise, holes, and open boundaries, making the construction of local triangulation extremely challenging, which makes the performance of \emph{Heat} and \emph{NManifold} degrade.
Conversely, our approach circumvents such challenges by leveraging learned geometric priors from the data, thereby surpassing the performance of existing methods.

\begin{table*}[t]
    \tablestyle{7pt}{1.2}
    \caption{Quantitative comparisons on Kinect v2 data. The MSE, the rate of MSE larger than 1.0 ($R_{MSE>1}$), and the average number of non-zero elements per row are reported. 
    }
    \begin{tabular}{cccccccccccccccc}
    \toprule
    \multirow{2}{*}{\#Num.} & \multicolumn{3}{c}{Graph\scriptsize{~\cite{Taubin1995}}} && \multicolumn{3}{c}{Heat\scriptsize{~\cite{Belkin2009}}} && \multicolumn{3}{c}{NManifold\scriptsize{~\cite{Sharp2020a}}} && \multicolumn{3}{c}{Ours} \\
    \cmidrule{2-4} \cmidrule{6-8} \cmidrule{10-12} \cmidrule{14-16}
     & MSE & Failure & Sparsity& & MSE & Failure & Sparsity& & MSE & Failure & Sparsity& & MSE & Failure & Sparsity  \\
    72 & 0.3120 & 11.53\% & 9.6& & 0.0264 & 0.47\% & 9.4 && 0.0246 & 0.33\% & 7.1 && 0.0034 & 0.02\% & 9.6 \\

    \bottomrule
    \end{tabular}
    \label{tab:real_world_scene_table}
\end{table*}

\subsection{Ablations and Discussions} \label{sec:ablation}

In this section, we discuss the key designs in our method and analyze the properties of our method.
All MSEs are calculated using the same rule as in ~\cref{evaluation_metrics} unless specifically mentioned.

\paragraph{Probe Functions}
The topological discrepancy between $L$ and $L_{gt}$ prevents direct supervision of $L$ using $L_{gt}$.
Thus, we adopt probe functions to observe the behavior of Laplacians to train the GNN.

We employ two types of spatial and spectral probe functions for training, as discussed in \cref{sec:probe}.
To verify the necessity of combining them, we train our network with only spatial probe functions, only spectral probe functions, and both.
During testing, we additionally introduce polynomial probe functions (described in ~\cref{evaluation_metrics}) to evaluate the generalization ability of our neural Laplacian operator on unseen functions.
The MSEs on the testing set are shown in the table as follows.

\begin{center}
    \tablestyle{8pt}{1.2}
    \vspace{2pt}
    \begin{tabular}{c|ccc}
                           & Spatial+Spectral & Spectral only & Spatial only \\\midrule
        Trigonometric      & 0.0086         & 0.0094     & 0.0086 \\
        Polynomial         & 0.0073         & 0.0098     & 0.0088 \\
        Eigenfunctions     & 0.0022         & 0.0018     & 0.0303 \\\midrule
        Total MSE          & 0.0049         & 0.0051     & 0.0210 \\
    \end{tabular}
    \vspace{2pt}
    \label{tab:probe_function}
\end{center}

For the overall MSEs in the last row, the resulting Laplacian operator performs significantly worse when trained with only spectral probe functions;
the MSEs are acceptable when trained with only spectral probe functions and are the best when trained with both.
Concerning unseen polynomial probe functions in the second row, the Laplacian operator trained with only spectral probe functions performs the worst;
the Laplacian operator works better when trained with two kinds of probe functions.

Crucially, spectral probe functions enable networks to learn the intrinsic geometry of thin structures.
These functions exhibit significant value differences across opposite faces, as illustrated in \cref{fig:spectral_probe_advantage}, where the green and yellow points, while spatially close, have distinct values.
The network will be strongly punished by the MSE loss if it fails to distinguish the two points.
Such a property enforces the network to learn to distinguish the two opposite surfaces of thin structures, which is usually difficult for traditional methods.

\paragraph{The $K$-Nearest Neighbor}
We investigate the impact of the number of nearest neighbors $K$ for constructing the KNN graph.
We vary $K$ to 4, 8, and 16 while maintaining all other training settings unchanged.
We set the batchsize to 3 for $K=16$ due to GPU memory constraints.
The MSEs on the testing set are as follows.
We also list the average number of non-zero elements in the Laplacian matrix per row (Sparsity).
\begin{center}
    \vspace{2pt}
    \tablestyle{12pt}{1.1}
    \begin{tabular}{c|ccc|c}
                   & $K=4$    & $K=8$       & $K=16$  & Mesh    \\  \midrule
    MSE            & 0.0066   & 0.0049      & 0.0048  &  -      \\
    Sparsity       & 6.76      & 9.8         & 18.3    & 7.0     \\
    \end{tabular}
    \vspace{2pt}
\end{center}
For $K=4$, the KNN graph is sparse, reducing its expressiveness and resulting in inferior performance.
For $K=16$, the KNN graph becomes one times denser, incurring higher computational costs.
Notably, the average degree of vertices in a triangular mesh is usually around 6.
Thus, in our experiments, we use $K=8$, resulting in a KNN graph with an average degree close to a regular triangular mesh.
This configuration enables our neural Laplacian operator to strike a balance between expressiveness and computational efficiency.

\paragraph{Graph Convolution}
We introduce a simple yet effective graph convolution operator in our network in ~\cref{eq:conv}, together with all one input signals in~\cref{sec:method:input_signal}.
The input signal prevents the network from knowing any information concerning the absolute coordinates of the points.
The graph convolution operator is designed to take local geometry properties, such as relative vertex position and edge length, into consideration. This is helpful for the network to understand local geometry structures.
The combination of these two designs makes our network only aware of the local geometry information, which is crucial for the Laplacian operator to be translation-invariant and scalable.
To further justify this design choice, we compare it with two other widely used graph convolution operators, GraphSAGE~\cite{Hamilton2017} and GATv2~\cite{Brody2022}.
However, feeding all-one input signal to GraphSAGE and GATv2 will make them fail to learn any geometric information, since they cannot learn realtive geometric relations between points during message passing process.
So we feed them with the absolute coordinates of the points as input signal. Note that this naturally breaks the scalibility and translation invariance of the network.
We maintain the training settings unchanged, except for reducing the batch size of GATv2 to 4 due to increased memory requirements.
The MSEs on the testing set are as follows:
\begin{center}
\vspace{2pt}
\tablestyle{16pt}{1.1}
\begin{tabular}{c|ccc}
        & GraphSAGE   & GAT      & Ours    \\ \midrule
    MSE & 0.0064      & 0.0063    & 0.0049   \\
\end{tabular}
\vspace{2pt}
\end{center}
The results verify the efficacy of our graph convolution operator compared to other popular designs for our task.

\begin{figure}[b]
    \centering
    \includegraphics[width=0.94\linewidth]{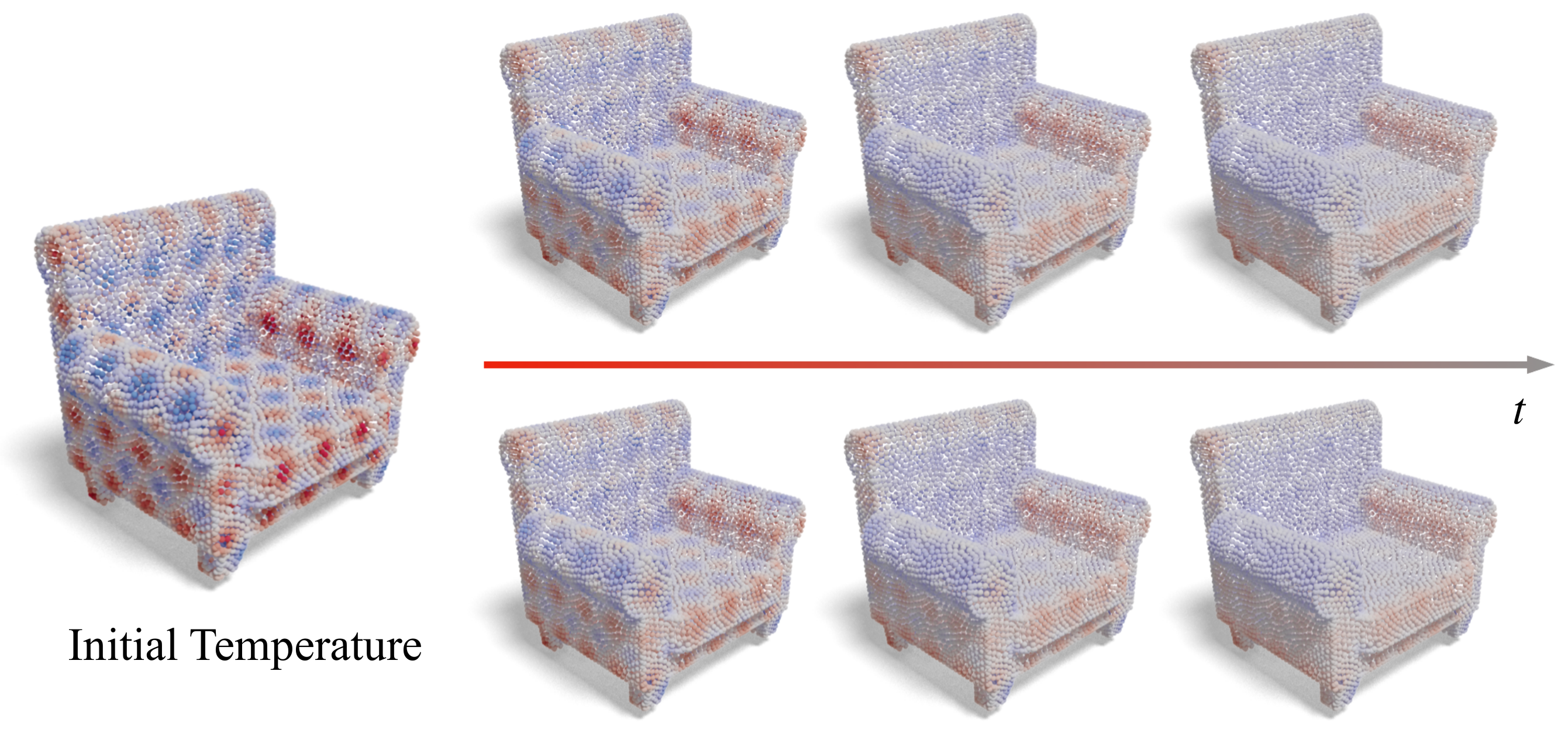}
    \caption{Heat diffusion process at different time steps.
    We use the explicit Euler method with a time step of 0.001, and visualize the heat distribution after 1000, 2000, and 3000 timesteps. The upper row is the result of our Laplacian operator, while the lower row is the result of the ground truth.}
    \label{fig:heat_diffusion}
\end{figure}

\paragraph{Properties of the Neural Laplacian Operator}
Here, we discuss various properties of our neural Laplacian operator (\textsc{NeLo}), laying the theoretical foundation for its application in geometry processing on point clouds~\cite{Wardetzky2007}.
\begin{itemize}[leftmargin=16pt,itemsep=2pt]
    \item[-] \emph{Symmetry}: For connectivity, KNN graph is made symmetric by adding reverse edges. For weights of edges, the symmetry of \textsc{NeLo} is guaranteed since the edge weight $w_{ij}$ equals $w_{ji}$ as defined in \cref{eq:edge}.
    \item[-] \emph{Locality}: \textsc{NeLo} relies on the KNN graph, ensuring its locality. Its sparsity (mean degree of vertices) is basically equivalent to that of a regular triangular mesh.
    \item[-] \emph{Non-negative weights}: The edge weight $w_{ij}$ is guaranteed to be non-negative since we append a ReLU to the MLP in \cref{eq:edge}.
    \item[-] \emph{Semi-definiteness}: Following the formulation in \cref{eq:laplacian}, we have $f^T L f = - \frac{1}{2} \sum_i \sum_j w_{ij} (f_i-f_j)^2$, where $f_i$ represents the $i$-th component of $f$. Consequently, \textsc{NeLo} is semi-definite since $w_{ij}$ is always non-negative.
    \item[-] \emph{Convergence}: We empirically observe that our algorithm exhibits a reasonable convergence property when densifying the point clouds, although we cannot guarantee it theoretically.  An example is shown in \cref{fig:limitation_convergence}.
\end{itemize}

\paragraph{\rev{Point Cloud Laplacian via Reconstruction}}
\rev{Given a point cloud, it is possible to first reconstruct a triangular mesh and then calculate the Laplacian of the mesh.
However, such an approach faces two obstacles. First, reconstructing a mesh from a sparse and non-oriented point cloud is also non-trivial and still an open problem. In contrast, we can solve the point cloud Laplacian problem in an end-to-end fashion.
Second, a large number of reconstruction methods does not guarantee that the original points will be interpolated in the final mesh and may introduce new vertices. This can cause issues if the Laplacian on the original point cloud is required.}

\subsection{Applications} \label{sec:application}

In this section, we apply several representative geometry processing algorithms that rely on the Laplacian operator of point clouds.
The objective is to showcase the new possibilities for point cloud processing enabled by our neural Laplacian operator.

\paragraph{Heat Diffusion}
The diffusion equation describes how thermal energy in a system diffuses over time.
It is a fundamental tool in physics-based simulations and other computer graphics applications.
The diffusion equation is defined as $\nabla^2 u = \partial u / \partial t$, which relies on the Laplacian operator.
The results of the diffusion process on an input point cloud are illustrated in \cref{fig:heat_diffusion}.

\begin{figure}[t]
    \centering
    \includegraphics[width=0.95\linewidth]{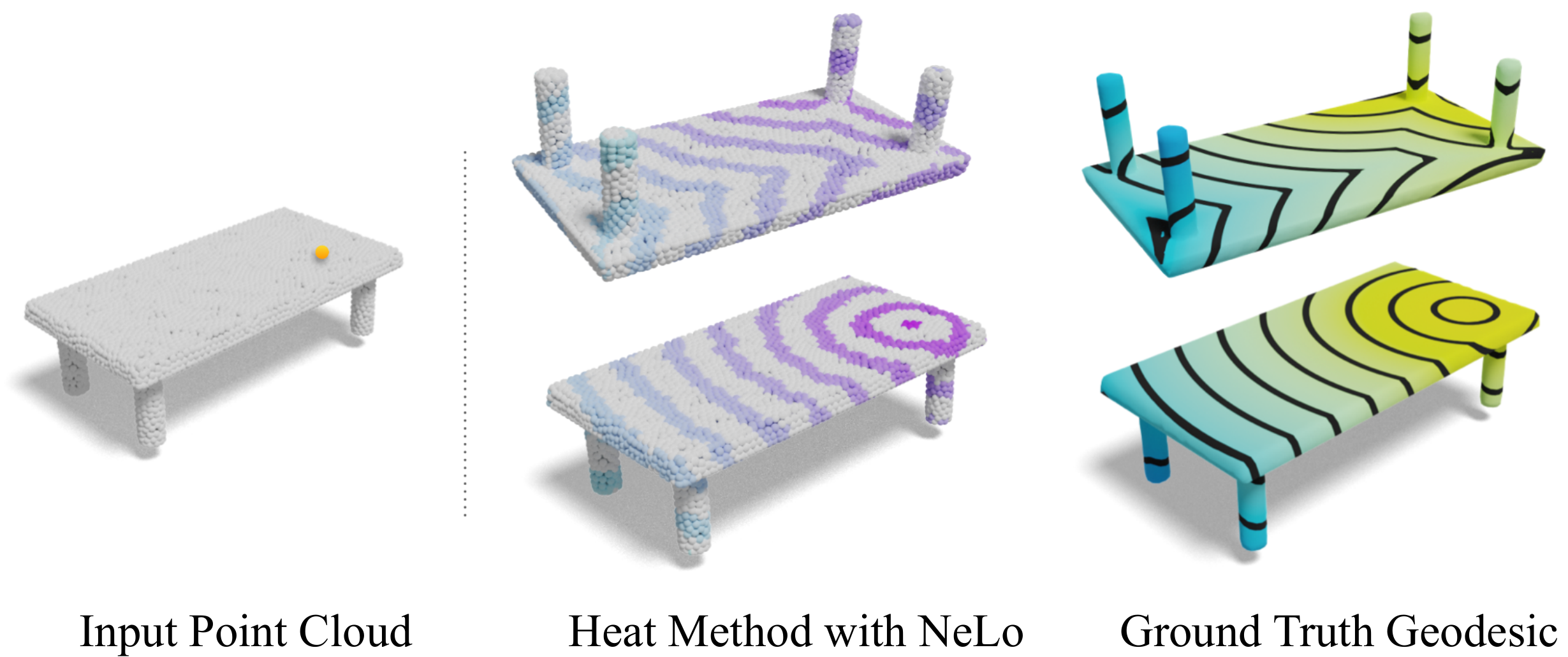}
    \caption{Geodesic distance.
    The source of the geodesic distance is annotated as the yellow spot.
    }
    \label{fig:geodesic_desk}
\end{figure}

\paragraph{Geodesic Distance}
The geodesic distance is the shortest distance between two points along the surface of a 3D shape.
This distance can be efficiently calculated by solving the heat equation and the Poisson equation~\cite{Crane2013a} on the shape which rely on the Laplacian operator.
We keep all other operators on the mesh unchanged but replace the Laplacian-Beltrami operator in the heat equation with our neural Laplacian operator.
The results are shown in \cref{fig:geodesic_desk}.
It can be seen that the geodesic distance is calculated accurately, both on the flat desktop and thin legs, and corresponds well to the ground truth mesh geodesic distance.

\begin{figure}[t]
    \centering
    \includegraphics[width=\linewidth]{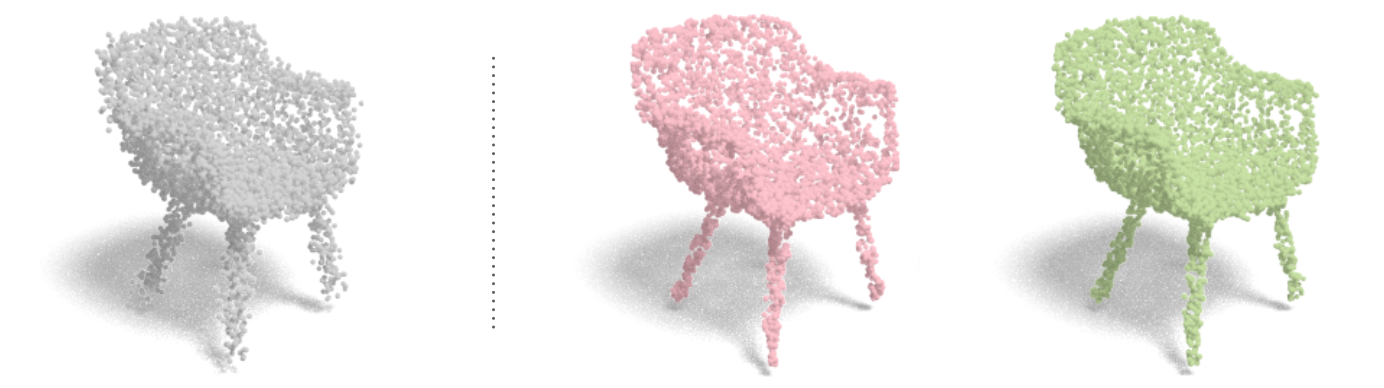}
    \caption{Laplacian smoothing.
    Given a noisy point cloud (left), we use our Laplacian matrix (middle) and ground truth Laplacian (right) to smooth the point cloud.
    }
    \label{fig:laplacian_smoothing}
    \vspace{-4pt}
\end{figure}

\paragraph{Laplacian Smoothing}
Laplacian smoothing~\cite{Desbrun1999} is commonly employed in 3D shape denoising and smoothing.
It iteratively adjusts the positions of vertices following the flow distribution of vertices provided by the Laplacian operator.
The results of Laplacian smoothing on a point cloud are presented in \cref{fig:laplacian_smoothing}.

\begin{figure}[t]
    \centering
    \includegraphics[width=\linewidth]{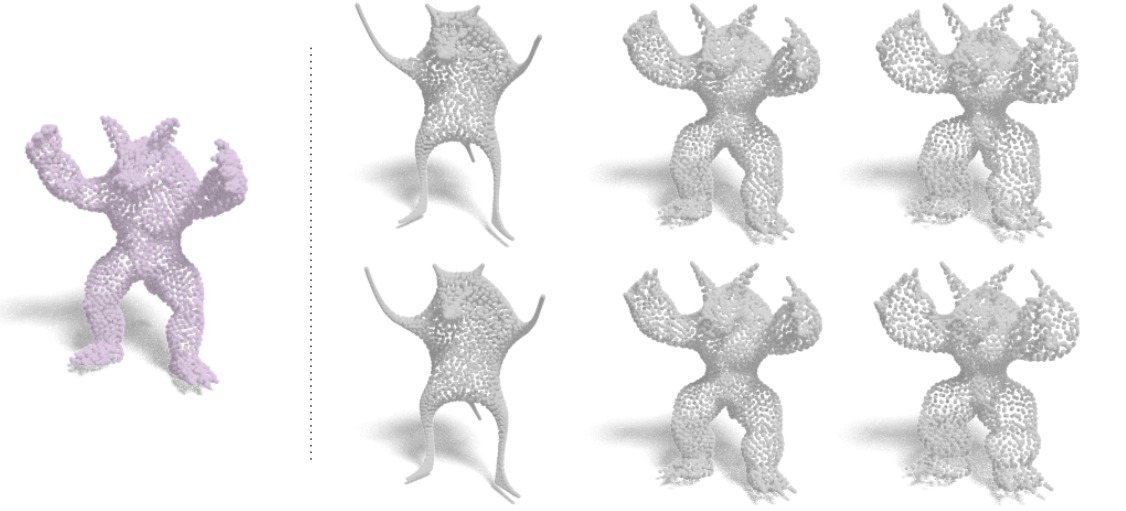}
    \caption{Spectral filtering.
    From left to right:
    The original point cloud;
    Low-pass filtering that retains the first 20 frequencies;
    High-pass filtering that reduce the first 20 frequencies to be 70\% of the original;
    Amplification that amplifies the first 20-200 frequencies by 2 times.
    The upper row is the result of our Laplacian operator, and the lower row is the result of the ground truth Laplacian operator.
    }
    \label{fig:spectral_filtering}
\end{figure}

\paragraph{Spectral Filtering}
The eigenvectors of the Laplace matrix can be utilized to extend the concept of the Fourier transform to discrete 2-manifold surfaces, with the eigenvalues representing the corresponding frequencies~\cite{Vallet2008}.
With our neural Laplacian operator, we can calculate eigenvectors and eigenvalues for the Laplace matrix of a point cloud, allowing the application of spectral filtering for point cloud smoothing or editing.
In \cref{fig:spectral_filtering}, we present the results of low-pass filtering and feature enhancement applied to a point cloud of the armadillo model.

\begin{figure}[t]
    \centering
    \includegraphics[width=0.98\linewidth]{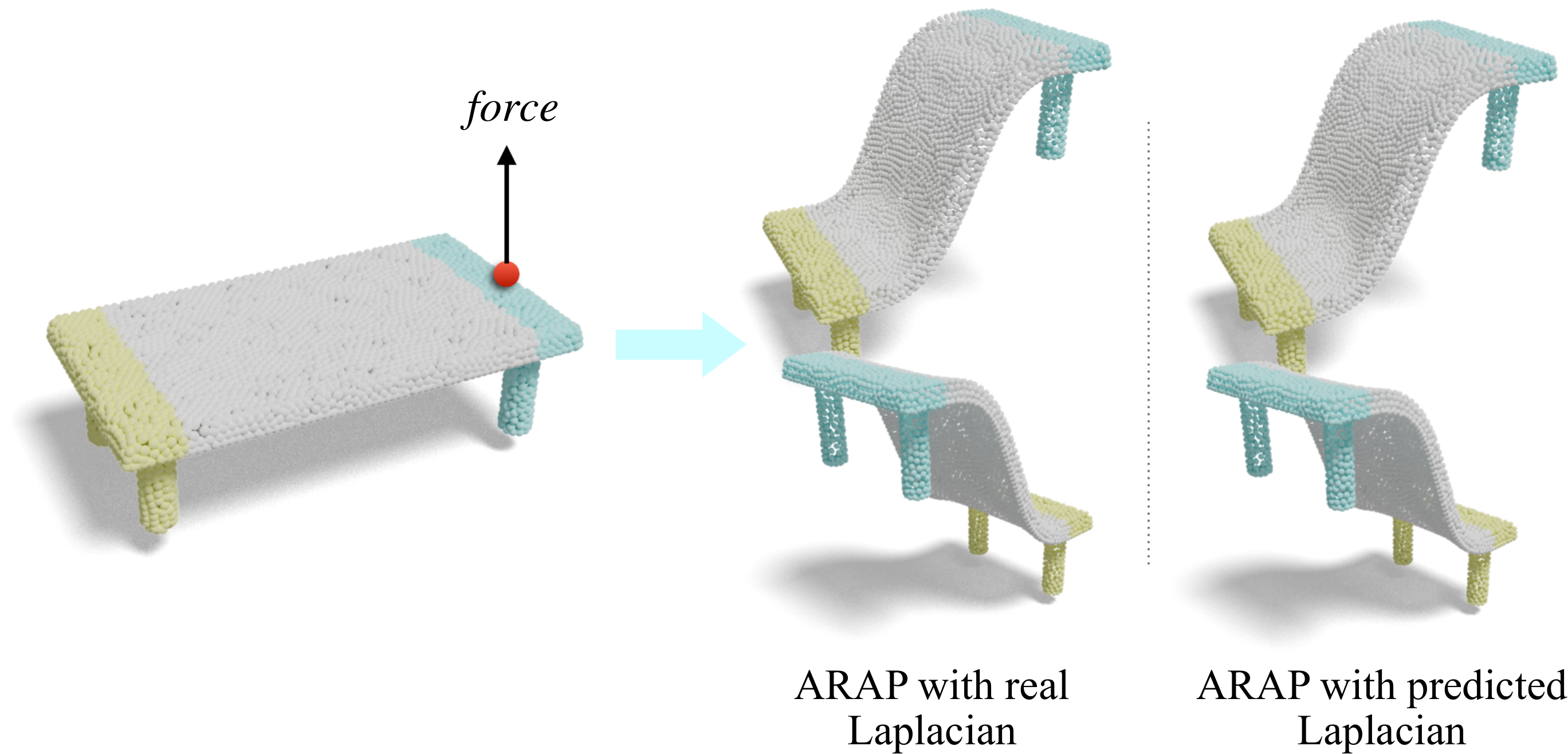}
    \caption{Deformation.
    The ARAP deformation is applied to the input point cloud on the left by fixing the yellow vertices and moving the cyan vertices to the target positions, producing the result on the right.
    }
    \label{fig:arap}
\end{figure}

\paragraph{Deformation}
The ``As Rigid As Possible'' (ARAP) deformation is a well-known method for 3D shape editing and manipulation~\cite{Sorkine2007}.
This method employs the Laplacian operator to enforce local rigidity constraints.
By solving linear systems derived from the Laplacian operator, an optimal deformation can be obtained that adheres to the constraints and preserves the structure of the original shape.
We apply the ARAP deformation to a point cloud utilizing our neural Laplacian operator, as illustrated in \cref{fig:arap}.

%% file: src/conclusion.tex
\section{Conclusion} \label{sec:conclusion}

In this paper, we propose a novel method to learn the Laplacian operator for point clouds using graph neural networks (GNNs).
The GNNs are trained to extract per-point features with a KNN graph as input, and the Laplacian operator is then computed from the features using MLPs.
We also propose a novel training scheme that leverages probe functions to address the issue of lacking ground-truth edge weights for supervising the learning of the Laplacian operator.
After training, our neural Laplacian operator behaves similarly to the ground-truth Laplacian operator on a set of probe functions.
The performance of our neural Laplacian operator significantly surpasses that of previous methods.
We further apply a series of geometry processing algorithms directly to point clouds with our neural Laplacian operator, demonstrating our method's potential in point cloud geometry processing.

The following outlines the limitations of our method and directions for future research.

\paragraph{Convergence Guarantee}
When the density of the input point cloud becomes infinitely large, our neural Laplacian operator is \emph{not} guaranteed to converge to the ground-truth Laplacian operator, which is one of our limitations.
As a learning-based method, the trained GNN performs the best on point clouds that are similar to the training set.
If the input point cloud becomes much denser or sparser than the training set,  the performance of our neural Laplacian operator may degrade.

\begin{figure}[t]
  \centering
  \includegraphics[width=0.87\linewidth]{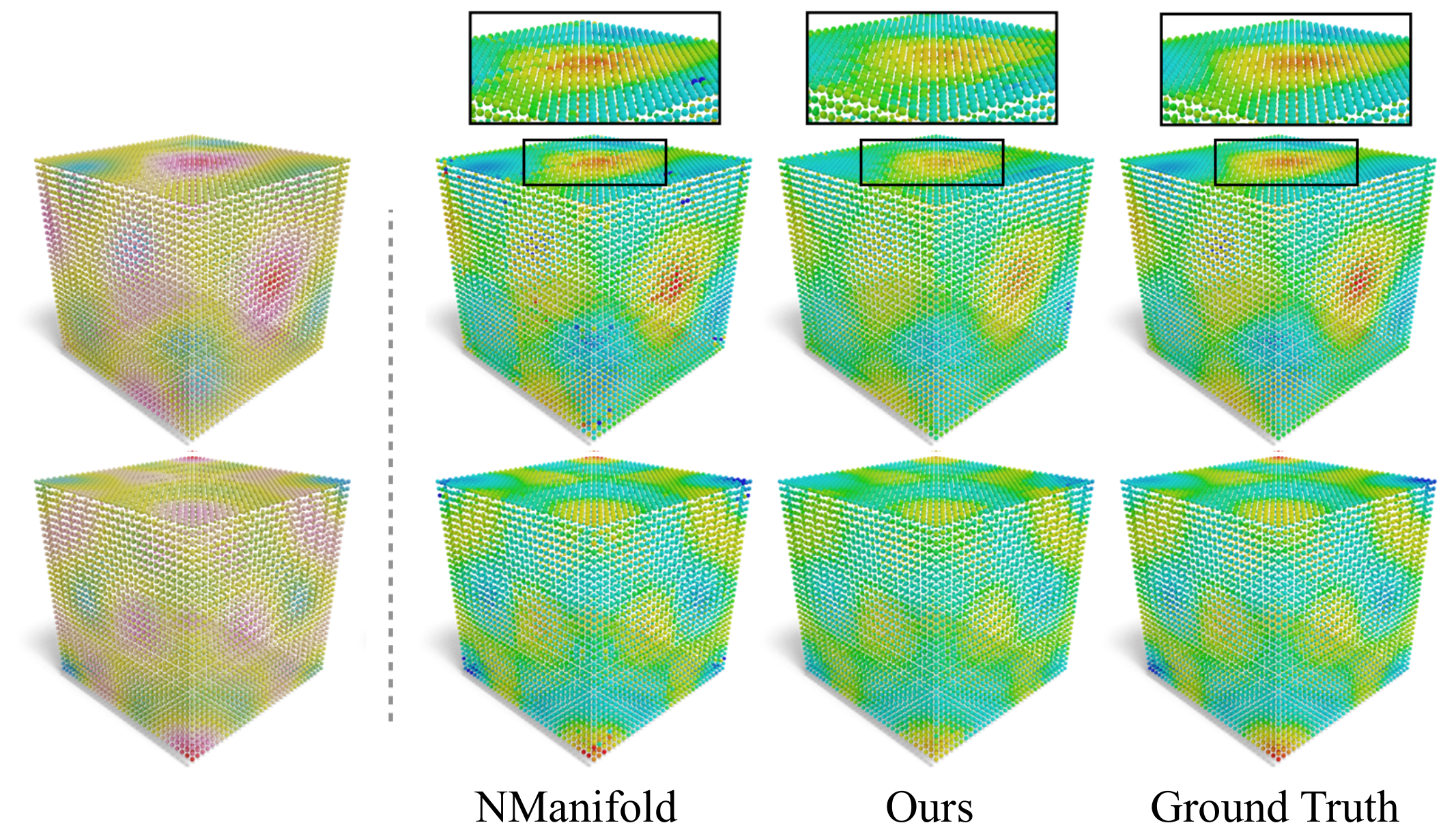}
  \caption{Results on a high-quality point cloud.
  We construct a point cloud of cube, on which points are uniformly distributed. On most regions of the cube, a triangle mesh can be easily reconstructed.
  We visualize the result of the ground truth Laplacian operator, NManifold~\cite{Sharp2020a}, and our method, respectively.
  }
  \label{fig:limitation_upper_bound}
\end{figure}

\paragraph{Accuracy Upper Bound}
Our network is trained to emulate the Laplacian-Beltrami operator defined on a mesh; thus, the accuracy of our neural Laplacian operator is bounded by it.
When the input point cloud is of high quality, allowing for the reconstruction of an accurate triangle mesh, previous triangulation-based algorithms may outperform ours.
In~\cref{fig:limitation_upper_bound}, we show the results for a point cloud uniformly sampled from a cube.
For such a clean and uniformly-sampled point cloud, a manifold mesh can be easily reconstructed using Delaunay triangulation.
It could be observed that, on flat regions (where the triangle mesh is well reconstructed), the result of NManifold is closer to the ground truth than our method, though at the corners of the cube it will have some artifacts.
To address this issue, it would be intriguing to integrate a classification mechanism that automatically assesses the quality of the input point cloud and selects our method when faced with challenges in reconstructing a high-quality triangle mesh from the input point cloud.

\begin{figure}[t]
  \centering
  \includegraphics[width=0.96\linewidth]{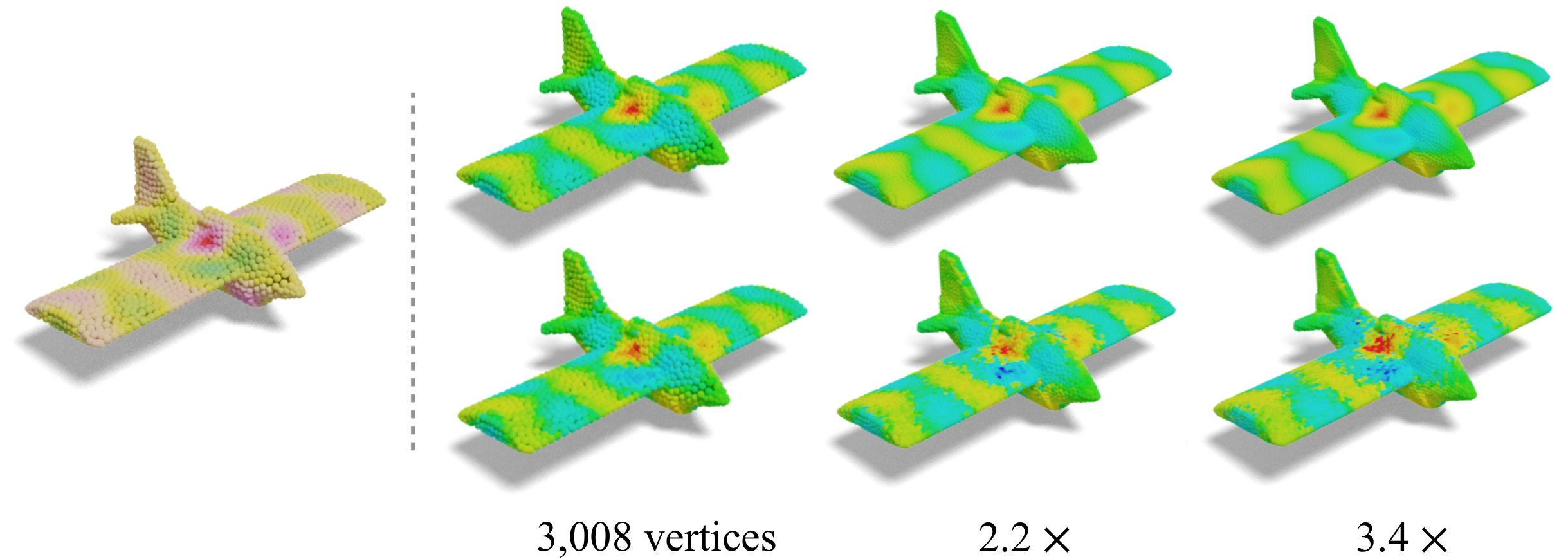}
  \caption{Convergence issue.
  On the left is a probe function.
  On the right are the results of our method and the ground truth Laplacian operator, with increasing point cloud density.
  The first row is the result of ground truth Laplacian, and the second row is the result of our method.
  }
  \label{fig:limitation_convergence}
\end{figure}

\paragraph{Intrinsic Delaunay Triangulations}
We use libigl~\cite{Libigl2018} to compute the Laplacian operator on the manifold mesh. However, negative-weight edges may occur in obtuse triangles.
This issue does not affect our current training data as they are processed to be good quality triangulation.
Sharp and Crane~\shortcite{Sharp2020a} proposed to leverage intrinsic Delaunay triangulations to define the Laplacian operator on meshes, ensuring good quality Laplacian on arbitrary meshes.
Employing such a method to compute the ground-truth Laplacian would allow the training on more diverse point clouds. \looseness=-1

\paragraph{Differential Operators}
This paper mainly focuses on learning the Laplacian operator for point clouds.
Our network and the training scheme are general and not limited to the Laplacian operator.
In the future, it is possible to explore other differential operators, such as the gradient operator and divergence operator for point clouds.
These operators would be useful for many more geometry processing algorithms on point clouds.

\paragraph{\rev{Advanced Networks}}
The proposed Laplacian is translation-invariant since we adopt the all-constant input signals.
However, it is not rotation-invariant.
Future works can try to substitute it with a different feature that is not sensitive to rotation, which will make the Laplacian rigid-invariant.